\documentclass[runningheads]{llncs}
\usepackage{graphicx}
\usepackage{comment}
\usepackage{amsmath,amssymb} % define this before the line numbering.
\usepackage{color}

\usepackage{multirow} 
\usepackage{tabularx, booktabs}
\usepackage[pagebackref=true,breaklinks=true,letterpaper=true,colorlinks,bookmarks=false]{hyperref}
\usepackage{arydshln}
\usepackage{enumitem}
\usepackage{dsfont}
\usepackage{subcaption}
\usepackage{breakcites}
\usepackage{algorithm}
\usepackage[noend]{algpseudocode}
\usepackage[dvipsnames]{xcolor}

\newcommand{\fig}[1]{Figure~\ref{#1}}
\newcommand{\sect}[1]{Section~\ref{#1}}
\newcommand{\tbl}[1]{Table~\ref{#1}}
\newcommand{\eqn}[1]{Equation~\ref{#1}}
\newcommand{\app}[1]{Appendix~\ref{#1}}
\newcommand{\alg}[1]{Algorithm~\ref{#1}}
\newcommand{\ignorethis}[1]{}

\newcommand{\xpar}[1]{\noindent\textbf{#1}\ \ }
\newif\ifsubmit

\submitfalse

\ifsubmit
\newcommand{\junyanz}[1]{}
\else
\newcommand{\junyanz}[1]{\textcolor{blue}{JY: #1}}
\fi

\newcommand{\comm}[1]{}

\newcommand{\red}[1]{\textcolor{red}{#1}}

\newcommand{\cyan}[1]{\textcolor{cyan}{#1}}
\newcommand{\cL}{\mathcal{L}}
\newcommand{\myparagraph}[1]{\vspace{.2cm} \noindent \textbf{#1}~}

\newcommand\blfootnote[1]{%
  \begingroup
  \renewcommand\thefootnote{}\footnote{#1}%
  \addtocounter{footnote}{-1}%
  \endgroup
}
\newcommand{\bc}{\mathbf{c}}
\newcommand{\bm}{\mathbf{m}}

\newcommand{\by}{\mathbf{y}}
\newcommand{\bz}{\mathbf{z}}

\newcommand{\bmu}{\mathbf{\mu}}
\newcommand{\cT}{\mathcal{T}}
\newcommand{\bSigma}{\mathbf{\Sigma}}

\newcommand{\cstar}{\mathbf{c^{*}}}
\newcommand{\zstar}{\mathbf{z^{*}}}

\DeclareMathOperator*{\argmin}{arg\,min}

\newif\ifeccv
\eccvfalse

\begin{document}
\pagestyle{headings}
\mainmatter
\def\ECCVSubNumber{2834}  % Insert your submission number here

\title{Transforming and Projecting Images into Class-conditional Generative Networks}

\begin{comment}
\titlerunning{ECCV-20 submission ID \ECCVSubNumber} 
\authorrunning{ECCV-20 submission ID \ECCVSubNumber} 
\author{Anonymous ECCV submission}
\institute{Paper ID \ECCVSubNumber}
\end{comment}
%******************

%\begin{comment}
\titlerunning{ }
\author{Minyoung Huh$^{12*}$ \quad 
Richard Zhang$^{2}$ \quad
Jun-Yan Zhu$^{2}$ \quad \\
Sylvain Paris$^{2}$ \quad
Aaron Hertzmann$^{2}$
}
\authorrunning{Huh, Zhang, Zhu, Paris, Hertzmann}
\institute{$^1$MIT CSAIL \qquad $^2$Adobe Research}

%\end{comment}
%******************
\maketitle

\begin{abstract}
We present a method for projecting an input image into the space of a class-conditional generative neural network. We propose a method that optimizes for transformation to counteract the model biases in generative neural networks. Specifically, we demonstrate that one can solve for image translation, scale, and global color transformation, during the projection optimization to address the object-center bias and color bias of a Generative Adversarial Network. This projection process poses a difficult optimization problem, and purely gradient-based optimizations fail to find good solutions. We describe a hybrid optimization strategy that finds good projections by estimating transformations and class parameters. We show the effectiveness of our method on real images and further demonstrate how the corresponding projections lead to better editability of these images. The project page and the code is available at \url{https://minyoungg.github.io/pix2latent}.
\end{abstract}

\blfootnote{*Work started during an internship at Adobe Research.}

\vspace{-0.3in}

\section{Introduction}
Deep generative models, particularly Generative Adversarial Networks (GANs)\\~\cite{goodfellow2014generative}, can create a diverse set of realistic images, with a number of controls for transforming the output, e.g., \cite{dcgan, bau2018gan,jahanian2019steerability,karras2019style,dcgan,bau2020rewriting,harkonen2020ganspace}. 
However, most of these methods apply only to synthetic images that are generated by GANs in the first place.  
In many real-world cases, a user would like to edit their own image.  One approach is to train a network for each separate image transformation. However, this would require a combinatorial explosion of training time and model parameters. 

Instead, a user could ``project'' their image to the manifold of images produced by the GAN, by searching for an appropriate latent code~\cite{zhu2016generative}. Then, any transformations available within the GAN could be applied to the user's image. This could allow a powerful range of editing operations within a relatively compact representation.
However, projection is a challenging problem. Previous methods have focused on class-specific models, for example, for objects~\cite{zhu2016generative},  faces~\cite{perarnau2016invertible,brock2017neural}, or specific scenes such as bedrooms and churches~\cite{bau2019semantic,bau2019seeing}. 
With the challenges in both optimization and generative model's limited capacity, we wish to find a generic method that can fit  \textit{real} images from diverse categories into the same generative model.

\begin{figure}[t!]
\vspace{-0.05in}
\includegraphics[width=1.0\linewidth]{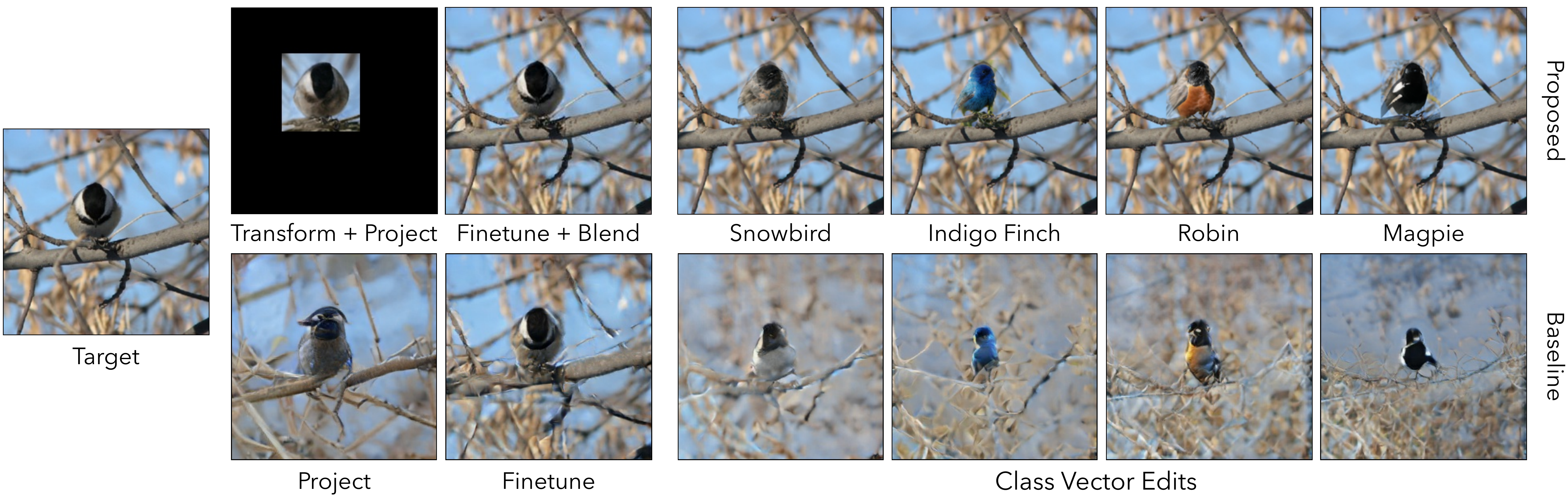}
\vspace{-0.1in}
\caption{
\small
Given a pre-trained BigGAN~\cite{biggan} and a target image (left), our method uses gradient-free BasinCMA to transform the image and find a latent vector to closely reconstruct the image. Our method (top) can better fit the input image, compared to the baseline (bottom), which does not model image transformation and uses gradient-based ADAM optimization. Finding an accurate solution to the inversion problem allows us to further fine-tune the model weights to match the target image without losing downstream editing capabilities. For example, our method allows for changing the class of the object (top row), compared to the baseline (bottom).
}
\vspace{-0.2in}
\label{fig:teaser}
\end{figure}

This paper proposes the first method for projecting images into class-conditional models. In particular, we focus on BigGAN \cite{biggan}. We address the main problems with these tasks, mainly, the challenges of optimization, object alignment, and class label estimation:
\vspace{0.03in}
\renewcommand\labelitemi{$\vcenter{\hbox{\tiny$\bullet$}}$}
\begin{itemize}[noitemsep,nolistsep,leftmargin=*]
    \item To help avoid local minima during the optimization process, we systematically study choices of both gradient-based and gradient-free optimizers and show Covariance Matrix Adaptation (CMA)~\cite{hansen2001cma} to be more effective than stand-alone gradient-based optimizers, such as L-BFGS~\cite{liu1989limited} and Adam~\cite{kingma2014adam}.
    \item To better fit a real image into the latent space, we account for the model's center bias by simultaneously estimating both spatial image transformation (translation,  scale, and color) and latent variable. Such a transformation can then be inverted back to the input image frame. Our simultaneous transformation and projection method largely expands the scope and diversity of the images that a GAN can reconstruct.
    \item Finally, we show that estimating and jointly optimizing the continuous embedding of the class variable leads to better projections. This ultimately leads to more expressive editing by harnessing the representation of the class-conditional generative model.

\end{itemize}

\vspace{0.03in}

We evaluate our method against various baselines on projecting real images from ImageNet. We quantitatively and qualitatively demonstrate that it is crucial to simultaneously estimate the correct transformation during the projection step.  Furthermore, we show that CMA, a non-parametric gradient-free optimization technique, significantly improves the robustness of the optimization and leads to better solutions. As shown in~\fig{fig:teaser}, our method allows us to fine-tune our model to recover the missing details without losing the editing capabilities of the generative model.

\section{Related Work}

\xpar{Image editing with generative models.}
Image editing tools allow a user to manipulate a photograph according to their goal while producing realistic visual content. Seminal work is often built on low-level visual properties, such as patch-based texture synthesis~\cite{efros1999texture,hertzmann2001image,efros2001image,barnes2009patchmatch}, gradient-domain image blending~\cite{perez2003poisson}, and image matting with locally affine color model~\cite{levin2007closed}. Different from previous hand-crafted low-level methods, several recent works~\cite{zhu2016generative,brock2017neural} proposed to build editing tools based on a deep generative model, with the hope that a generative model can capture high-level information about the image manifold. 

\ifeccv
    Many prior works have investigated using trained generative models as a tool to edit images~\cite{zhu2016generative, brock2017neural,bau2019semantic,bau2018gan}. The same image prior from deep generative models has also been used in face editing, image inpainting, colorization, and deblurring prior~\cite{perarnau2016invertible,yeh2017semantic,asim2018blind,gu2020image,shen2020interpreting}. 
    Unlike these works that focuses on single-class and fixated image, our method presents a new ways of embedding an image into a class-conditional generative model, which allows the same GAN to be applied to many more “in-the-wild” scenarios. 
\else
    For example, iGAN~\cite{zhu2016generative} proposes to reconstruct and edit a real image using GANs. 
    The method first projects a real photo onto a latent vector using a hybrid method of encoder-based initialization and per-image optimization.  It then modifies the latent vector using various editing tools such as color, sketch, and warping brushes and generates the final image accordingly. Later, Neural Photo Editing~\cite{brock2017neural} proposes to edit a face photo using VAE-GANs~\cite{larsen2016autoencoding}. The same image prior from deep generative models has also been used in face editing, image inpainting, colorization,  and deblurring~\cite{perarnau2016invertible,yeh2017semantic,asim2018blind,gu2020image,shen2020interpreting}. Recently, GANPaint~\cite{bau2019semantic} proposes to change the semantics of an input image by first projecting an image into GANs, then fine-tuning the GANs to reproducing the details,  and finally modifying the intermediate activations based on user inputs~\cite{bau2018gan}. However, all the above systems focus on a single object category. Our work presents new ways of embedding an image into a class-conditional generative model, which allows the same GAN to be applied to many more “in-the-wild” scenarios. 

\fi

\begin{figure}[t!]
    \centering
    \includegraphics[width=1.0\linewidth]{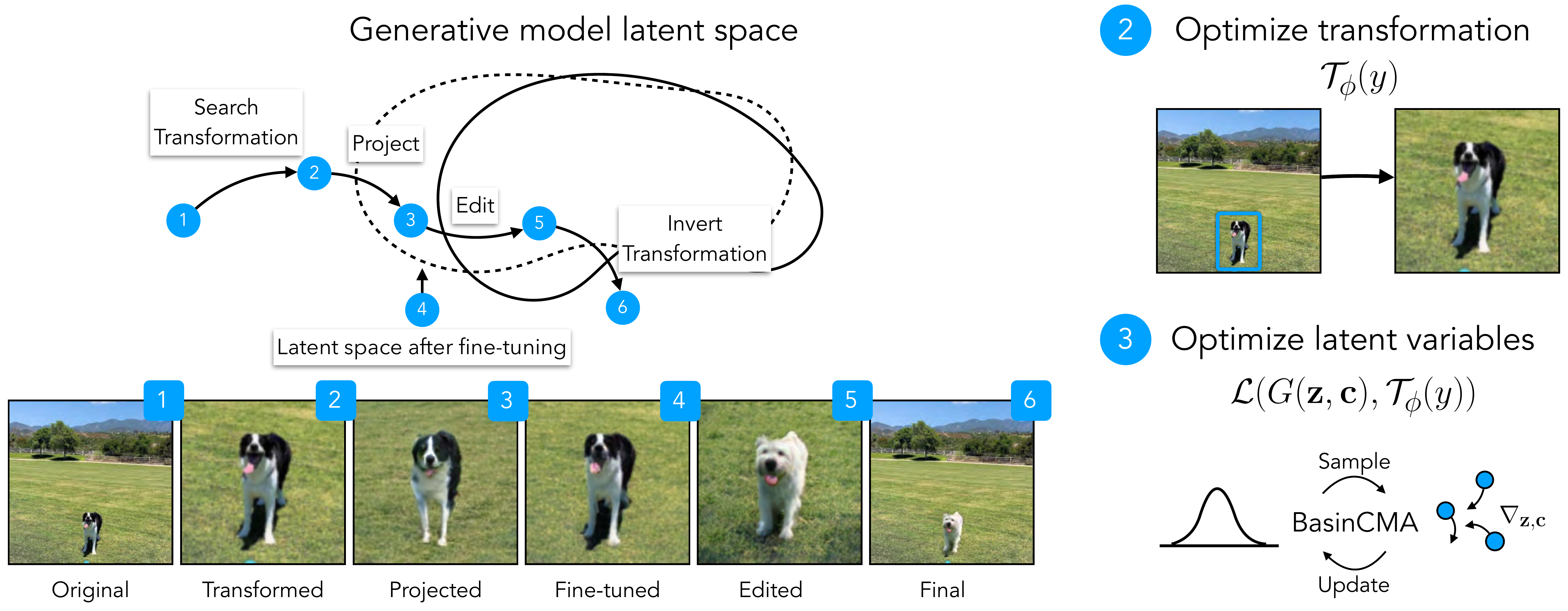}
    \vspace{-.2in}
    \caption{\small \textbf{Overview:} Our method first searches for a transformation to apply to the input target image. We then solve for the latent vector that closely resembles the object in the target image, using our proposed optimization method, also referred to as ``projection''. The generative model can then be further fine-tuned to reconstruct the missing details that the original model could not generate. Finally, we can edit the image by altering the latent code or the class vector (e.g., changing the border collie to a west highland white terrier), and invert and blend the edited image back into the original image.}
    \vspace{-.2in}
    \label{fig:model}
\end{figure}

\xpar{Inverting networks.} Our work is closely related to methods for inverting pre-trained networks. Earlier work proposes to invert CNN classifiers and intermediate features for visualizing recognition networks~\cite{mahendran2015understanding,dosovitskiy2016inverting,olah2017feature,olah2018building}. 
More recently, researchers adopted the above methods to invert generative models. 
The common techniques include: (1) Optimization-based methods: they find the latent vector that can closely reconstruct the input image using gradient-based method (e.g., ADAM, LBFGS)~\cite{zhu2016generative,brock2017neural,lipton2017precise,yeh2017semantic,shahhegde,rajlibresler,creswell2018inverting} or MCMC~\cite{fangschwing}, (2) Encoder-based methods: they learn an encoder to directly predict the latent vector given a real image~\cite{donahue2016adversarial,dumoulin2016adversarially,zhu2016generative,perarnau2016invertible,brock2017neural,donahue2019large}, (3) Hybrid methods~\cite{zhu2016generative,bau2019semantic,bau2019seeing}: they use the encoder to initialize the latent vector and then solve the optimization problem. 
\ifeccv
    Although the optimized latent vector roughly approximates the real input image, many important visual details are missing in the reconstruction~\cite{bau2019semantic}. To address the issue, GANPaint~\cite{bau2019semantic} generates residual features to adapt to the individual image. Image2StyleGAN~\cite{abdal2019image2stylegan} optimizes StyleGAN's intermediate representation rather than the input latent vector. Unfortunately, the above techniques still cannot handle images in many scenarios due to the limited model capacity~\cite{bau2019seeing}, the lack of generalization ability~\cite{abdal2019image2stylegan}, and their single-class assumption. As noted by prior work~\cite{abdal2019image2stylegan}, the reconstruction quality severely degrades under simple image transformation, and translation has been found to cause most of the damage. Compared to prior work, we consider two new aspects in the reconstruction pipeline: image transformation and class vector. Together, these two aspects significantly expand the diversity of the images that we can reconstruct and edit. 
\else
    By using the ground truth class label, Neural Collage~\cite{suzuki2018spatially} turn a BigGAN into a single-class GAN and adopts single-class image projection~\cite{zhu2016generative}. Although the optimized latent vector roughly approximates the real input image, many important visual details are missing in the reconstruction~\cite{bau2019semantic}. To address the issue, GANPaint~\cite{bau2019semantic} gently fine-tunes the pre-trained GAN to adapt itself to the individual image. Image2StyleGAN~\cite{abdal2019image2stylegan} optimizes StyleGAN's intermediate representation rather than the input latent vector. Unfortunately, the above techniques still cannot handle images in many scenarios due to the limited model capacity~\cite{bau2019seeing}, the lack of generalization ability~\cite{abdal2019image2stylegan}, and their single-class assumption. As noted by prior work~\cite{abdal2019image2stylegan}, the reconstruction quality severely degrades under simple image transformation, and translation has been found to cause most of the damage. Compared to prior work, we consider two new aspects in the reconstruction pipeline: image transformation and class vector. Together, these two aspects significantly expand the diversity of the images that we can reconstruct and edit. 

\fi
\section{Image projection methods}

We aim to project an image into a class-conditional generative model (e.g., BigGAN~\cite{biggan}) for the purposes of downstream editing. We first introduce the basic objective function that we slowly build upon. 
Next, since BigGAN is an object-centric model for most classes, we infer an object mask from the input image and focus on fitting the pixels inside the mask.

Furthermore, to better fit our desired image into the generative model, we propose to optimize for various image transformation (scale, translation, and color) to be applied to the target image.
Lastly, we explain how we optimize the aforementioned objective loss function.

\subsection{Basic Loss Function}

\myparagraph{Class-conditional generative model} A class-conditional generative network can synthesize an image $\hat{\mathbf{y}}\in\mathds{R}^{H\times W\times 3}$, given a latent code $\bz\in\mathds{R}^{Z}$ that models intra-class variations and a one-hot class-conditioning vector $\tilde{\bc}\in\Delta^{\mathcal{C}}$ to choose over $C$ classes. We focus on the $256\times256$ BigGAN model~\cite{biggan} specifically, where $Z=128$ and $C=1{,}\,000$ ImageNet classes.

The BigGAN architecture first maps the one-hot $\tilde{\bc}$ into a continuous vector $\bc\in\mathds{R}^{128}$ with a linear layer $\mathbf{W}\in \mathds{R}^{128\times 1000}$, before injecting into the main network $G_{\theta}$, with learned parameters $\theta$.

\begin{equation}
    \hat{\mathbf{y}} = G_{\theta}(\bz,\bc) = G_{\theta}(\bz,\mathbf{W}\tilde{\bc}).
\end{equation}

\noindent Here, a choice must be made whether to optimize over the discrete $\tilde{\bc}$ or continuous $\bc$. As optimizing a discrete class vector is non-trivial, we optimize over the continuous embedding.

\myparagraph{Optimization setup.}
Given a target image $\by$, we would like to find a $\zstar$ and $\cstar$ that generates the image.
\begin{equation}
    \zstar,\cstar = \argmin_{\bz,\bc} \cL(G_{\theta}(\bz, \bc), \by)\qquad \text{s.t. } C(\bz) \le C_\mathrm{max}.
\end{equation}
During training, the latent code is sampled from a multivariate Gaussian \sloppy \mbox{$\bz \sim \mathcal{N}(\textbf{0}, \textbf{I})$}. Interestingly, recent methods~\cite{biggan,kingma2018glow} find that restricting the distribution at \emph{test time} produces higher-quality samples.
We follow this and constrain our search space to match the sampling distribution from Brock et al.~\cite{biggan}.
Specifically, we use \mbox{$C(\bz)=||\bz||_\infty$} and $C_\mathrm{max}=2$.
During optimization, elements of $\bz$ that fall outside the threshold are clamped to $+2$, if positive, or $-2$, if negative.
Allowing larger values of $\bz$ produces better fits but compromises editing ability.

\myparagraph{Loss function.} The loss function $\cL$ attempts to capture how close the approximate solution is to the target. A loss function that perfectly corresponds to human perceptual similarity is a longstanding open research problem~\cite{wang2004image}, and evaluating the difference solely on a per-pixel basis leads to blurry results~\cite{zhang2016colorful}. Distances in the feature space of a pre-trained CNN correspond more closely with human perception ~\cite{johnson2016perceptual,dosovitskiy2016generating,gatys2016image,zhang2018perceptual}. We use the LPIPS metric~\cite{zhang2018perceptual}, which calibrates a pre-trained model using human perceptual judgments. Here, we define our basic loss function, which combines per-pixel $\ell_1$ and LPIPS.

\begin{equation}\label{eq:inv_w_per}
\cL_\text{basic}(\by, \hat{\by}) =
    \frac{1}{H W} \lVert \hat{\by} - \by \rVert_1 + \beta \mathcal{L}_\text{LPIPS}(\hat{\by}, \by).
\end{equation}
In preliminary experiments, we tried various loss combinations and found $\beta=10$ to work well. We now expand upon this loss function by leveraging object mask information. 

\subsection{Object Localization}
\label{sec:obj_masking}

Real images are often more complex than the ones generated by BigGAN. For example, objects may be off-centered and partially occluded, or multiple objects appear in an image. Moreover, it is possible that the object in the image can be approximated by GANs but not the background.  

Accordingly, we focus on fitting a single foreground object in an image and develop a loss funciton to emphasize foreground pixels. We automatically produce a foreground rectangular mask $\mathbf{m}\in[0,1]^{H\times W\times 1}$ using the bounding box of an object detector~\cite{he2017mask}. Here, we opt for bounding boxes for simplicity, but one could consider using segmentation mask, saliency maps, user-provided masks, etc. The foreground and background values within mask $\mathbf{m}$ are set to $1$ and $0.3$, respectively. We adjust the objective function to spatially weigh the loss:

\begin{equation}\label{eq:masked_inv_w_per}
\cL_\text{mask}(\by, \hat{\by}, \bm) =
  \frac{1}{M}
    % \frac{1}{\sum_{hw} \bm}
   \lVert \bm \odot (\hat{\by} - \by) \rVert_1 + \beta \mathcal{L}_\text{mLPIPS}(\hat{\by}, \by, \textbf{m}), 
\end{equation}
where normalization parameter $M=\lVert \bm \rVert_1$ and $\odot$ represents element-wise multiplication across the spatial dimensions. Given a mask of all foreground (all ones), the objective function is equivalent to \eqn{eq:inv_w_per}. We calculate the masked version of the perceptual loss $\mathcal{L}_\text{mLPIPS}(\hat{\by}, \by, \textbf{m})$ by bilinearly downsampling the mask at the resolution of the intermediate spatial feature maps within the perceptual loss. The details are described in~Appendix~\red{B}.%\ref{app:mLPIPS}.
With the provided mask, we now explore how one can optimize for image transformation to better fit the object in the image.

\begin{figure}[t]
\vspace{-0.05in}
\begin{minipage}[b]{.48\textwidth}
\centering
\includegraphics[width=1.0\linewidth]{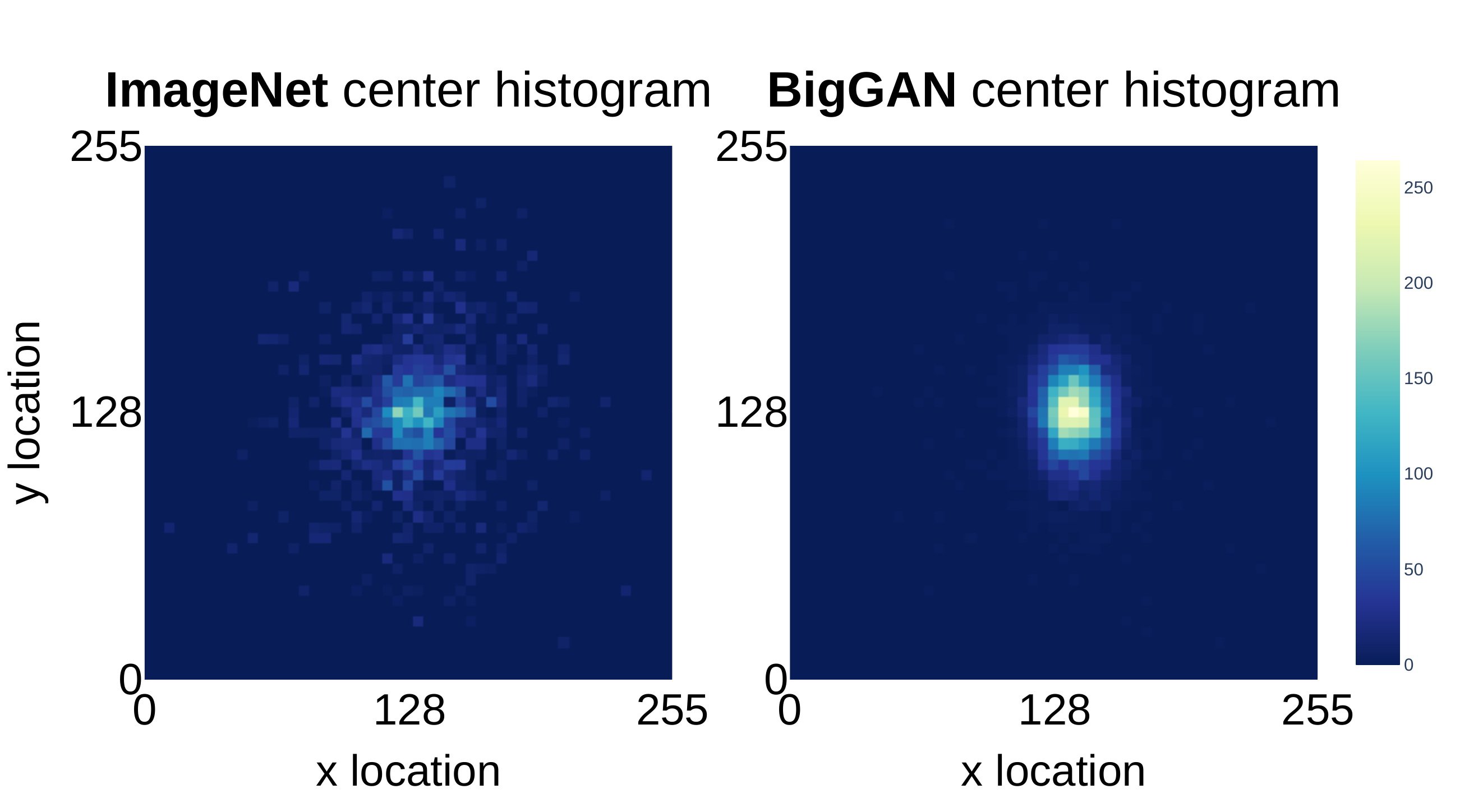}
\vspace{-0.3in}
\caption{\small \textbf{Object center comparison:} We use an object detector to compute the histogram of object locations. Note that ImageNet ({\bf left}) is biased towards the center but exhibits a long-tail. BigGAN ({\bf right}) is further biased towards center.}
\label{fig:comp_center}
\end{minipage}
\hfill
\begin{minipage}[b]{.5\textwidth}
\centering
\includegraphics[width=1.0\linewidth]{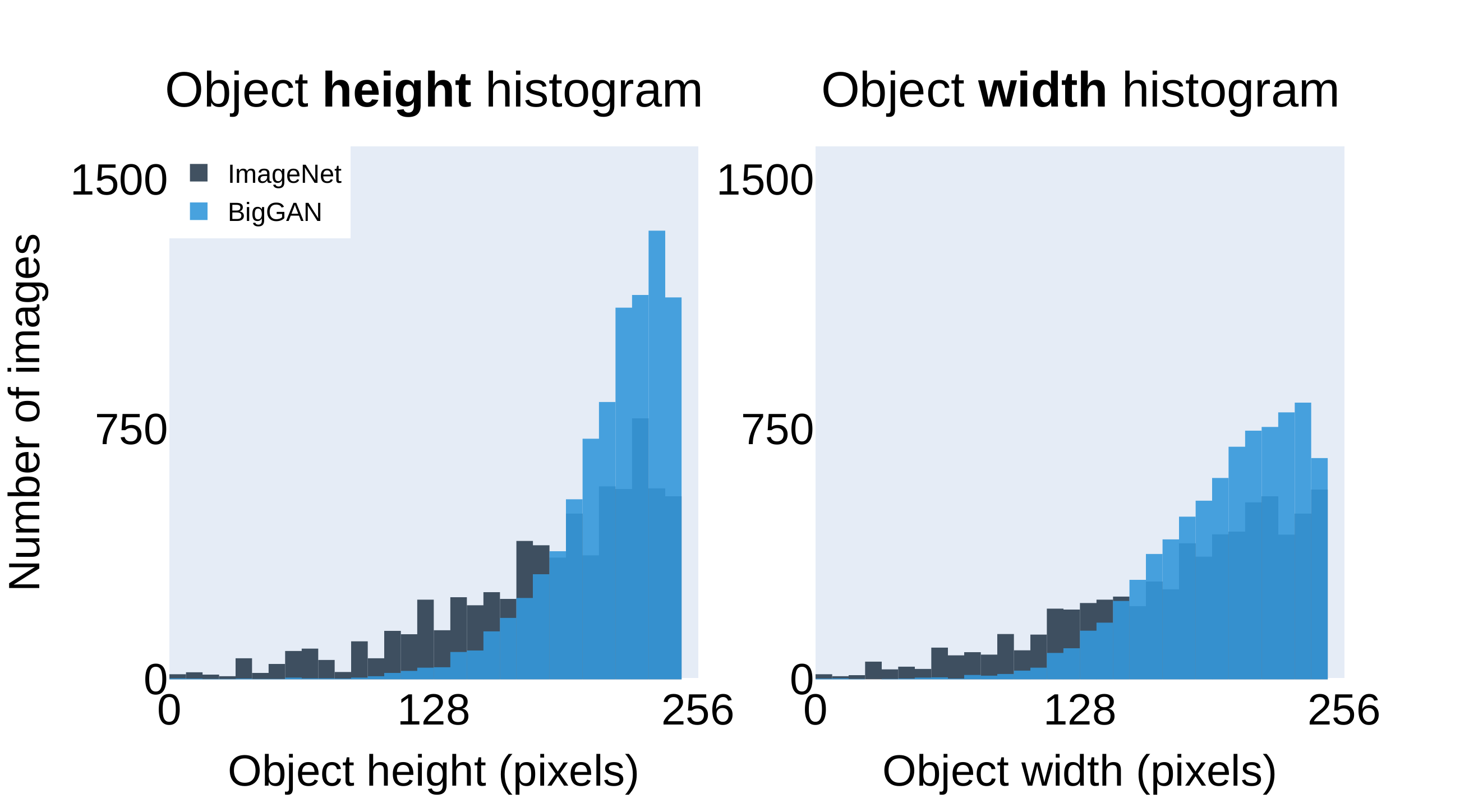}
\vspace{-0.3in}
\caption{\small \textbf{Object size comparison:} We use an object detector to compute the distribution of object widths ({\bf left}) and heights ({\bf right}). Note that ImageNet (black) has a long-tail, whereas the BigGAN  (blue) accentuates the mode.}
\label{fig:comp_size}
\end{minipage}
\vspace{-0.2in}
\end{figure}

\subsection{Transformation Model and Loss}
\label{sec:stats}

Generative models may exhibit biases for two reasons: (a) inherited biases from the training distribution and (b) bias introduced by mode collapse~\cite{goodfellow2016nips}, where the generative model only captures a portion of the distribution. We mitigate two types of biases, \textit{spatial} and \textit{color} during image reconstruction process. A concurrent work~\cite{anirudh2020mimicgan} proposes that surrogate network with spatial transformers could be used to partially alleviates this aforementioned distributional shift at test-time.

\myparagraph{Studying spatial biases.} To study spatial bias, we first use a pre-trained object detector, MaskRCNN~\cite{he2017mask}, over 10,000 real and generated images to compute the statistics of object locations. We show the statistics regarding the center locations and object sizes in Figures~\ref{fig:comp_center} and~\ref{fig:comp_size}, respectively.

Figure~\ref{fig:comp_center} (left) demonstrates that ImageNet images exhibit clear center bias over the location of objects, albeit with a long tail. While the BigGAN learns to mimic this distribution, it further accentuates the bias~\cite{bau2019seeing,jahanian2019steerability}, largely forgoing the long tail to generate high-quality samples in the middle of the image. In Figure~\ref{fig:comp_size}, we see similar trends with object height and width. Abdal et al.~\cite{abdal2019image2stylegan} noted that the quality of image reconstruction degrades given a simple translation in the target image. Motivated by this, we propose to incorporate spatial alignment in the inversion process.

\myparagraph{Searching over spatial alignments.} 
We propose to transform the generated image using $\cT^\text{spatial}_{\psi}(\cdot)$, which shifts and scales the image using parameters $\psi=[s_x,s_y,t_x,t_y]$. The parameters $\psi$ are used to generate a sampling grid which in turn is used by a grid-sampler to construct a new transformed image~\cite{jaderberg2015stn}. The corresponding inverse parameters are $\psi^{-1}=\big[\frac{1}{s_x},\frac{1}{s_y},-\frac{t_x}{s_x},-\frac{t_y}{s_y}\big]$.

Transforming the generated image allows for more flexibility in the optimization. For example, if $G$ can perfectly generate the target image, but at different scales or at off-centered locations, this framework allows it to do so.

\myparagraph{Searching over color transformations.} Furthermore, we 
show that the same framework allows us to search over color transformations $\cT^\text{color}_\gamma(\cdot)$. We experimented with various color transformations such as hue, brightness, gamma, saturation, contrast, and found brightness and contrast to work the best. Specifically, we optimize for brightness, which is parameterized by scalar $\gamma$ with inverse value $\gamma^{-1}=-\gamma$. If the generator can perfectly generate the target image, but slightly darker or brighter, this allows a learned brightness transformation to compensate for the difference.

\comm{
    \begin{align}
        \cL_\text{inv}(\by, \phi) = \frac{1}{MC} \lVert \bm \odot (\by - \cT_{\phi^{-1}} (\cT_{\phi}(\by))) \rVert_{2}.
     \end{align}
}

\myparagraph{Final objective.} Let transformation function $\cT_{\phi} = \cT^{\text{spatial}}_\psi \circ \cT^{\text{color}}_\gamma$ be a composition of spatial and color transformation functions, where transformation parameters $\phi$ is a concatenation of spatial and color parameters $\psi, \gamma$, respectively. The inverse function is $\cT_{\phi^{-1}}$. Our final optimization objective function, with consideration for (a) the foreground object and (b) spatial and color biases, is to minimize the following loss:

\begin{align}
    \argmin_{\bz,\bc,\phi} ~& \cL_\text{mask}(\cT_{\phi^{-1}}(G_{\theta}(\bz, \bc)), \by, \bm) \hspace{3mm}
    \mathrm{s.t.}~C(\bz) \le C_\mathrm{max} % \nonumber
    \label{eqn:transform_and_project}
\end{align}

Our optimization algorithm, described next, has a mix of gradient-free and gradient-based updates.  Alternatively, instead of inverse transforming the \textit{generated} image, we can transform the \textit{target} and mask images during gradient-based updates and compute the following loss: $\cL_\text{mask}(G_{\theta}(\bz, \bc), \cT_{\phi}(\by), \cT_{\phi}(\bm))$. We will discuss when to use each variant in the next section. 

\subsection{Optimization Algorithms}
\label{sec:optimize}

Unfortunately, the objective function is highly non-convex. Gradient-based optimization, as used in previous inversion methods, frequently fall into poor local minima. Bau et al.~\cite{bau2019seeing} note that recent large-scale GAN models~\cite{progressive_GAN,karras2019style} are significantly harder to invert due to a large number of layers, compared to earlier models~\cite{dcgan}. Thus, formulating an optimizer that reliably finds good solutions is a significant challenge.
We evaluate our method against various baselines and ablations in \sect{sec:results}. Given the input image $\by$ and foreground rectangular mask $\bm$ (which is automatically computed), we present the following algorithm.
\begin{figure*}[t!]
    \centering
    \includegraphics[width=1.0\linewidth]{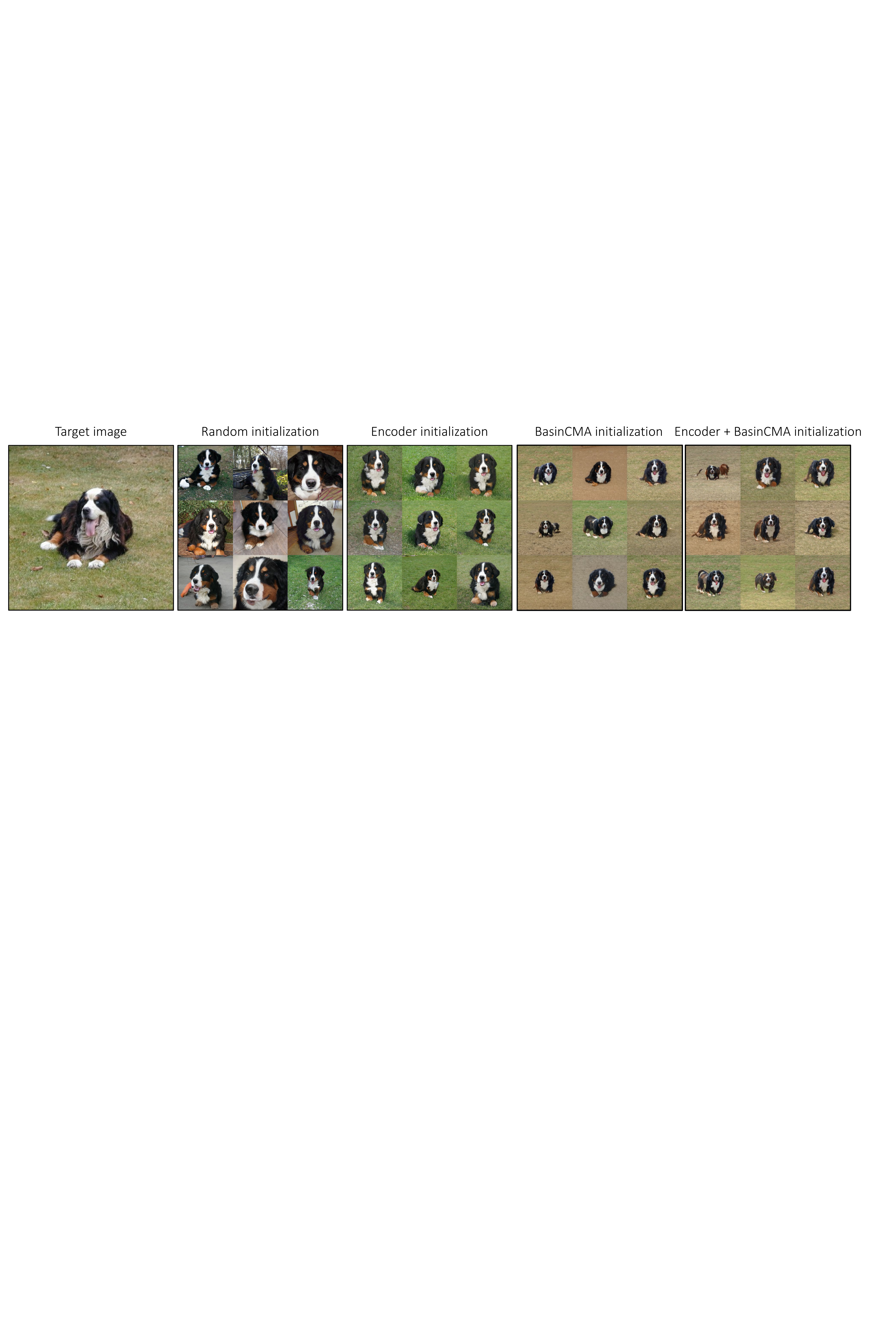}
    \vspace{-.1in}
    \caption{\small \textbf{Initialization from various methods:} We show samples drawn from different methods, before the final gradient descent optimization. In ``random initialization'', seeds are drawn from the normal distribution; the results show higher variation. For the ``encoder initialization'', we use a trained encoder network to predict the latent vector and apply a minor perturbation. Our method uses CMA to find a good starting distribution. For ``Encoder+BasinCMA'', we initialize CMA with the output of the encoder. The results are more consistent and better reconstruct the target image.
    }
    \vspace{-.2in}
    \label{fig:inits}
\end{figure*}

\myparagraph{Class and transform initialization}
We first predict the class of the image with a pre-trained ResNeXt101 classifier~\cite{xie2016@resnext} and multiply it by $\mathbf{W}$ to obtain our initial class vector $\bc_0$. 

Next, we initialize the spatial transformation vector $\psi_0=[s_{x_0}, s_{y_0}, t_{y_0}, t_{x_0}]$ such that the foreground object is well-aligned with the statistics of the BigGAN model.
As visualized in Figures~\ref{fig:comp_center} and~\ref{fig:comp_size}, $(\bar{h}, \bar{w})=(137,127)$ is the center of BigGAN-generated objects and $(\bar{y}, \bar{x})=(213,210)$ is the mode of object sizes. We define ($h_{\bm},w_{\bm}$) to be the height and width and ($y_{\bm},x_{\bm}$) to be center of the masked region. We initialize scale factors as $s_{y_0}=s_{x_0}=\max \big(\frac{h_{\bm}}{\bar{h}}, \frac{w_{\bm}}{\bar{w}}\big)$ and translations as $(t_{y_0},t_{x_0})= \big(\frac{\bar{y}-y_{\bm}}{2}, \frac{\bar{x}-x_{\bm}}{2}\big)$. Finally, initial brightness transformation parameter is initialized as $\gamma_0=1$.

\myparagraph{Choice of optimizer.} We find the choice of optimizer critical and that BasinCMA~\cite{bau2019seeing} provides better results than previously used optimizers for the GAN inversion problem.
Previous work~\cite{zhu2016generative,abdal2019image2stylegan} has exclusively used gradient-based optimization, such as LBFGS~\cite{liu1989limited} and ADAM~\cite{kingma2014adam}. However, such methods are prone to obtaining poor results due to local minima, requiring the use of multiple random initial seeds.
Covariance Matrix Adaptation (CMA)~\cite{hansen2001cma}, a \textit{gradient-free} optimizer, finds better solutions than gradient-based methods. CMA maintains a Gaussian distribution in parameter space $\bz\sim \mathcal{N}(\bmu,\bSigma)$. At each iteration, $N$ samples are drawn, and the Gaussian is updated using the loss. The details of this update are described in Hansen and Ostermeier~\cite{hansen2001cma}.
A weakness of CMA is that when it nears a solution, it is slow to refine results, as it does not use gradients.
To address this, we use a variant, BasinCMA~\cite{wampler2009basincma}, that alternates between CMA updates and ADAM optimization, where CMA distribution is updated after taking $M$ gradient steps.

Next, we describe the optimization procedure between the transformation parameters $\phi$ and latent variables $\bz, \bc$.

\begin{figure}[t!]
\vspace{-0.3in}
\begin{center}
    \scalebox{0.85}{
    \begin{minipage}{1.0\linewidth}
        \begin{algorithm}[H]
        \hspace*{\algorithmicindent} \textbf{Input:} Image $\by$, initial class vector $\bc_0$, mask $\bm$\\
        \hspace*{\algorithmicindent} \textbf{Output:} Transformation parameter $\phi^*$, latent variable $\bz^*$, class vector $\bc^*$
        \vspace{-0.05in}
        \newcommand{\algrule}[1][.2pt]{\par\vskip.5\baselineskip\hrule height #1\par\vskip.5\baselineskip}
        \caption{Transformation-aware projection algorithm}
        \begin{algorithmic}[1]
        \algrule
        \State \textcolor{MidnightBlue}{\texttt{\# Optimize for transformation} $\phi$}
        \State Initialize $(\mu_{\phi}, \Sigma_{\phi}) \leftarrow (\phi_0, 0.1 \cdot \textbf{I})$ \Comment{$\phi_0$ precomputed in \sect{sec:stats}}
        \For{{\rm \textbf{n}} iterations}
            \State $\phi_{1:N} \sim \text{SampleCMA}(\mu_{\phi}, \Sigma_{\phi})$ \Comment{Draw $N$ samples of $\phi$} 
            \State $\bz_{1:N} \sim \mathcal{N}(\textbf{0}, \textbf{I})$, reset $\bc_{1:N} \leftarrow \bc_0$ \Comment{Reinitialize $\bz$ and $\bc$}
            \For{{\rm \textbf{m}} iterations}
                \For{$i \leftarrow 1 {\rm \; \textbf{to} \;} N$}  \Comment{This loop is batched}
                    \State $g_i \leftarrow \cL_\text{mask}(G_{\theta}(\bz_{i}, \bc_{i}),  \cT_{\phi_{i}}(\by),\cT_{\phi_{i}}(\bm)) $
                \State $(\bz_i, \bc_i) \leftarrow (\bz_i, \bc_i) - \eta \cdot \nabla_{\bz, \bc} \;g_i$ \Comment{Update each sample $\bz$, $\bc$} 
                \EndFor
            \EndFor
            \State $g^{inv}_{1:N} \leftarrow \cL_\text{mask}(\cT_{\phi_{1:N}^{-1}}(G_{\theta}(\bz_{1:N}, \bc_{1:N})), \by, \bm) $\Comment{Recompute loss with inverse}
            \State $\mu_{\phi}, \phi_{1:N} \leftarrow \text{UpdateCMA}(\phi_{1:N},  g^{inv}_{1:N}, \mu_{\phi}, \Sigma_{\phi})$ \Comment{\sect{sec:optimize}}
        \EndFor
        \State Set $\phi^* \leftarrow \mu_{\phi}$ 
        \State \textcolor{MidnightBlue}{\texttt{\# Optimize for latent variables} $\bz, \bc$}
        \State Initialize ($\mu_{\bz}, \Sigma_{\bz}) \leftarrow (\mathbf{0}, \textbf{I})$
        \For{{\rm \textbf{p}} iterations}
            \State $\bz_{1:M} \sim \text{SampleCMA}(\mu_{\bz}, \Sigma_{\bz})$, reset $\bc_{1:M} \leftarrow \bc_0$ \Comment{Draw M samples of $\bz$}
            \For{{\rm \textbf{q}} iterations}
                \For{$i \leftarrow 1 {\rm \; \textbf{to} \;} M$} \Comment{This loop is batched}
                    \State $g_i \leftarrow \cL_\text{mask}(G_{\theta}(\bz_i, \bc_i), \cT_{\phi^{*}}(\by),\cT_{\phi^{*}}(\bm))$
                    \State $(\bz_i, \bc_i) \leftarrow (\bz_i, \bc_i)  - \nabla_{\bz, \bc} \; g_i $ 
                \EndFor
            \EndFor
            \State $g^{inv}_{1:N} \leftarrow \cL_\text{mask}(\cT_{\phi_{1:N}^{-1}}(G_{\theta}(\bz_{1:N}, \bc_{1:N})), \by, \bm) $\Comment{Recompute loss with inverse}
            \State $\mu_{\bz}, \Sigma_{\bz} \leftarrow \text{UpdateCMA}_{\bz}(\bz_{1:M},  g^{inv}_{1:M}, \mu_{\bz}, \Sigma_{\bz})$ \Comment{\sect{sec:optimize}}
        \EndFor
        \State Set $\bz^*, \bc^* \leftarrow \argmin_{\bz, \bc}(g_{1:M})$ \Comment{Choose the best $\bz, \bc$}
        %\EndFunction
        \end{algorithmic}
        \label{alg:projection}
        \end{algorithm}
    \end{minipage}}
\end{center}
\vspace{-0.2in}
\end{figure}

\myparagraph{Choice of Loss function} In~\eqn{eqn:transform_and_project}, we described two variants of our optimization objective. Ideally, we would like to optimize the 
former variant $\cL_\text{mask}(\cT_{\phi^{-1}}(G_{\theta}(\bz, \bc)), \by, \bm)$ such that the target image $\by$ is consistent through-out optimization; and we do so for all CMA updates. However for gradient optimization, we found that back-propagating through a grid-sampler to hurt performance, especially for small objects. A potential reason is that when shrinking a generated image, the grid-sampling operation sparsely samples the image. Without low-pass filtering, this produces a noisy and aliased result~\cite{gonzalez92imageproc,oppenheim99dtsp}. Therefore, for gradient-based optimization, we optimize the latter version $\cL_\text{mask}(G_{\theta}(\bz, \bc), \cT_{\phi}(\by), \cT_{\phi}(\bm))$.

\myparagraph{Two-stage approach.} Historically, searching over spatial transformations with reconstruction loss as guidance has proven to be a difficult task in computer vision~\cite{baker2004lucas}. We find this to be the case in our application as well, and that joint optimization over the transformation $\phi$, and variables $\bz, \bc$ is unstable. We use a two-stage approach, as shown in~\alg{alg:projection}, where we first search for $\phi^{*}$ and use $\phi^{*}$ to optimize for $\bz^{*}$ and $\bc^*$. In both stages, a gradient-free CMA outer loop maintains a distribution over the variable of interest in that stage. In the inner loop, ADAM is used to quickly find the local optimum over latent variables $\bz, \bc$. 

To optimize for the transformation parameter, we initialize CMA distribution for $\phi$. The mean $\mu_{\phi}$ is initialized with pre-computed statistics $\phi_0$, and $\Sigma_{\phi}$ is set to $0.1 \cdot \mathbf{I}$~(Alg.~\ref{alg:projection}, line~2). A set of transformations $\phi_{1:N}$ is drawn from CMA, and latent variables $\bz_{1:N}$ are randomly initialized~(Alg.~\ref{alg:projection}, line~4--5). To evaluate the sampled transformation, we take gradient updates w.r.t. $\bz_{1:N}, \bc_{1:N}$ for $m=30$ iterations~(Alg.~\ref{alg:projection}, line~6--9). This inner loop can be interpreted as quickly assessing the viability of a given spatial transform. The final samples of $\bz_{1:N}, \bc_{1:N}, \phi_{1:N}$ are used to compute the loss for the CMA update~(Alg.~\ref{alg:projection}, line~10--11). This procedure is repeated for $n=30$ iterations, and the final transformation $\phi^*$ is set to the mean of the current estimate of CMA~(Alg.~\ref{alg:projection}, line~12). 

After solving for the transformation $\phi^*$, a similar procedure is used to optimize for $\bz$. We initialize CMA distribution for $\bz$ with $\mu_{\bz} = \mathbf{0}$ and $\Sigma_{\bz}=\mathbf{I}$~(Alg.~\ref{alg:projection}, line~$14$). $M$ samples of $\bz_{1:M}$ are drawn from the CMA distribution and $\bc_{1:M}$ is set to the initial predicted class vector~(Alg.~\ref{alg:projection}, line~16). The drawn samples are evaluated by taking $q=30$ gradient updates w.r.t $\bz_{1:M}$ and $\bc_{1:M}$~(Alg.~\ref{alg:projection}, line~17--20). The optimized samples are used to compute the loss for the CMA update~(Alg.~\ref{alg:projection}, line~21--22). This procedure is repeated for $p=30$ iterations. On the final iteration, we take $300$ gradient updates instead to obtain the final solution $\bz, \bc$~(Alg.~\ref{alg:projection}, line~23).

\ifeccv
\else
    \begin{figure*}[t!]
    \centering
    \includegraphics[width=1.0\linewidth]{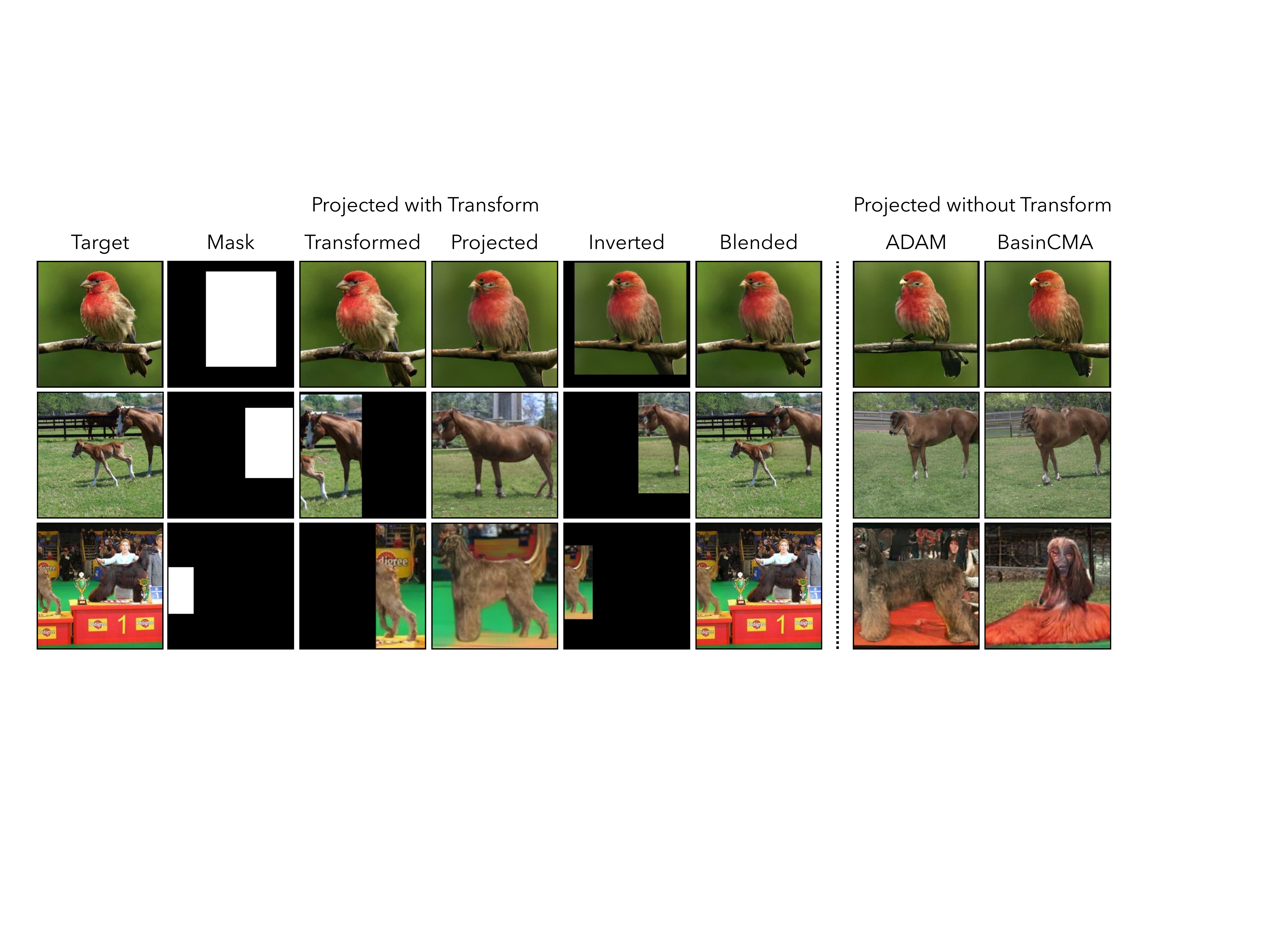}
    \vspace{-.1in}
    \caption{\small \textbf{Transformation search:} Our method optimizes for a  geometric and color transformation, such that the transformed image is more easily projected into the latent space. Transforming the image causes some pixels to be missing. To address this, we invert the results back to the original coordinates and poisson blend with the background after optimization. Results are not-finetuned.}
    \label{fig:transform}
    \vspace{-.2in}
\end{figure*}

\fi

\subsection{Fine-tuning} 

So far, we have located an approximate match within a generative model. We hypothesize that if a high-quality match is found, fine-tuning to fit the image will preserve the editability of the generative model. On the contrary, if a poor match is found, the fine-tuning will corrupt the network and result in low-quality images after editing. Next, we describe this fine-tuning process.

To synthesize the missing details that the generator could not produce, we wish to fine-tune our model after solving for the latent vector $\bz$, the class vector $\bc$, and transformation parameters $\phi$. Unlike previous work~\cite{bau2019semantic}, which proposed to produce the residual features using a small, auxiliary network, we update the weights of the original GAN directly. This allows us to perform edits that spatially deform the image. 
After obtaining the values for $\phi, \bz, \bc$ in our projection step, we fine-tune the weights of the generative model.
During fine-tuning, the full objective function is:

\begin{equation}
\begin{split}
\argmin_{\bz, \bc, \phi,\theta}
 \;\; \cL_\text{mask}(\cT_{\phi^{-1}}(G_{\theta}(\bz, \bc)), \by, \bm) 
 \; + \;  \lambda \lVert \theta - \theta_0 \rVert_2 \quad
\text{s.t. \;} C(\bz) \le C_\mathrm{max} \label{eqn:final}
\end{split}
\end{equation}

We put an $\ell_2$-regularization on the weights, such that the fine-tuned weights do not deviate too much from the original weights $\theta_0$. In doing so, we can prevent overfitting and preserve the generative model's ability to edit the final image. We use $\lambda = 10^{3}$ for our results with fine-tuning.

\begin{figure*}[t!]
        \centering
        \includegraphics[width=1.0\linewidth]{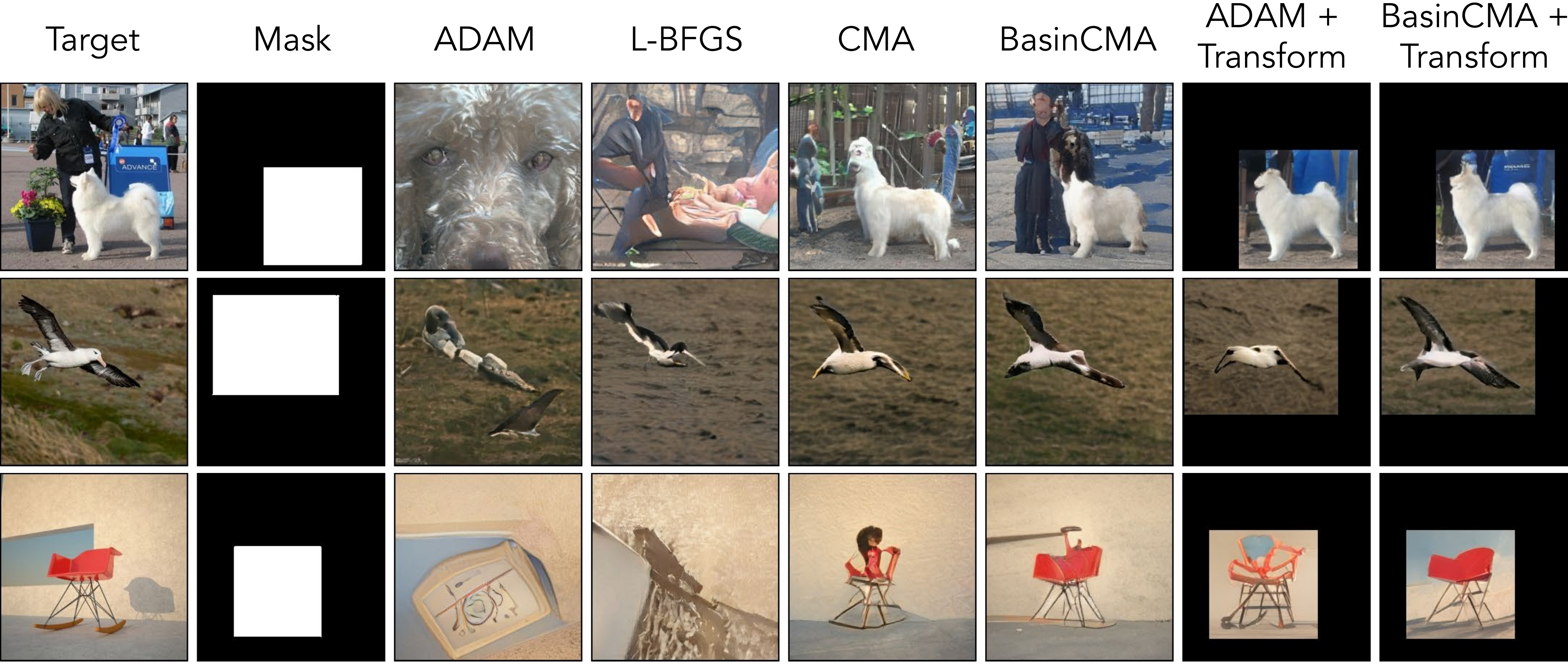}
        \vspace{-.23in}
        \caption{\small \textbf{ImageNet comparisons:} Comparison across various methods on inverting ImageNet images without fine-tuning. A rectangular mask centered around the object of interest is provided for all methods using MaskRCNN~\cite{he2017mask}. The losses are weighted by the mask. BasinCMA+Transform is our full method. 
        \vspace{-.1in}
        }
        \label{fig:comparisons}
\vspace{-0.05in}
\end{figure*}
\begin{figure*}[t!]
    \centering
    \includegraphics[width=1.0\linewidth]{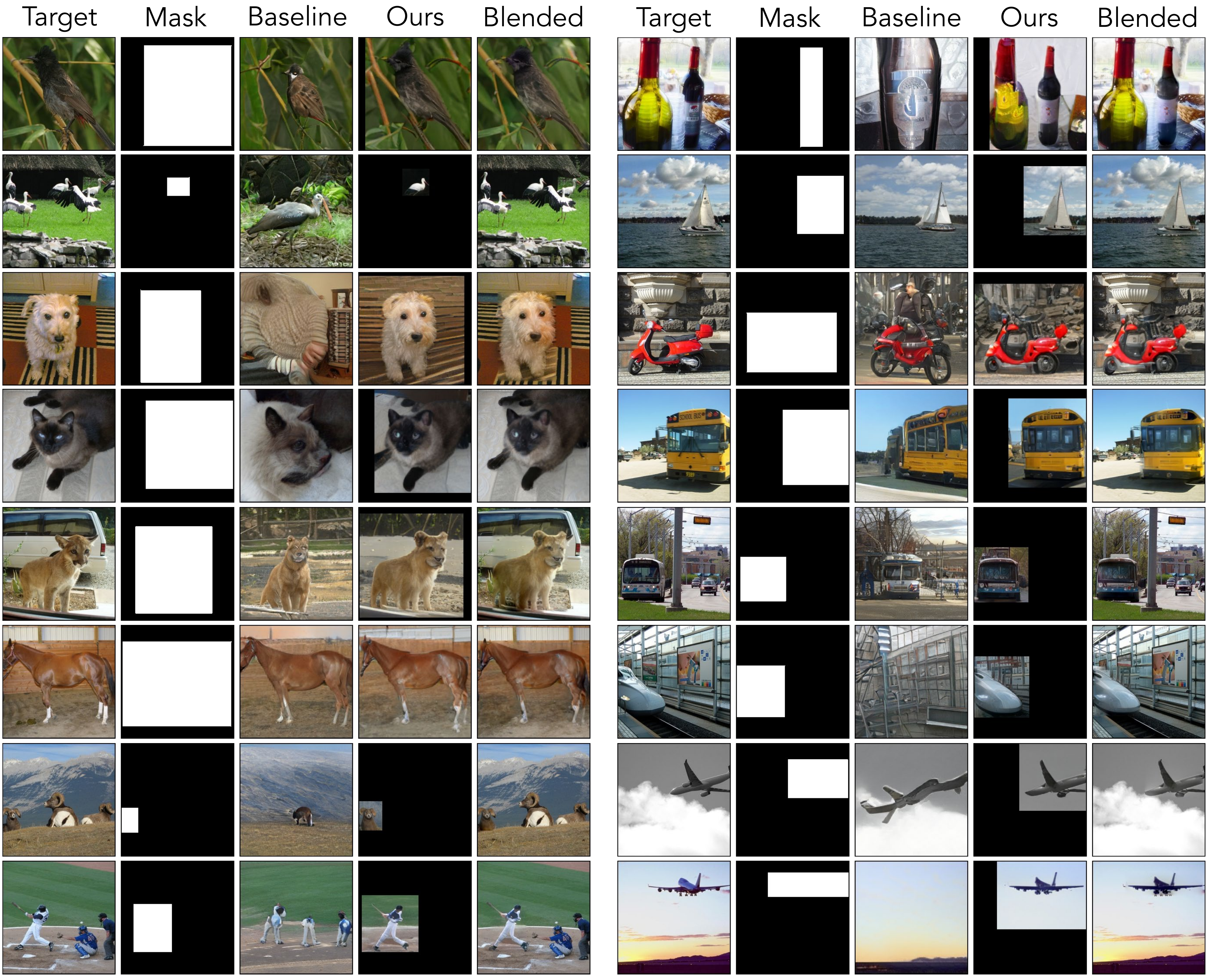}
    \vspace{-.23in}
    \caption{\small \textbf{ImageNet results:} Results using our final method without fine-tuning. The final method uses BasinCMA as well as spatial and color transformation. Our generated results are inverted back for visualization. We also provide the ADAM baseline along with the blended result using Poisson blending~\cite{perez2003poisson}. }
    \label{fig:results}
\vspace{-0.2in}
\end{figure*}

\section{Results}
\label{sec:results}

We demonstrate results on images from ImageNet~\cite{deng2009imagenet}, compare against baselines and ablations, examine cases that BigGAN cannot generate, and show failure cases. We further demonstrate the validity of our method on out-of-distribution data such as COCO and conduct perceptual studies on the edited images.

\label{sec:comparisons}

The ImageNet dataset consists of $1.3$ million images with $1{,}\,000$ classes. We construct a test set by using PASCAL~\cite{everingham2015pascalvoc} classes as super-classes. There are a total of $229$ classes from ImageNet that map to $16$ out of $20$ classes in PASCAL. We select $10$ images at random from each super-class to construct a dataset of $160$ images. We run off-the-shelf Mask-RCNN~\cite{he2017mask} and take the highest activating class to generate the detection boxes. We use the same bounding box for all baselines, and the optimization hyper-parameters are tuned on a separate set of ImageNet images. 

\setlength{\tabcolsep}{3.5pt}
\begin{table}[t]
\begin{center}
\resizebox{0.95\linewidth}{!}{
\begin{tabular}{cccc cccc cccc} 
    \toprule
    \multicolumn{4}{c}{\bf Method} & \multicolumn{4}{c}{\textbf{Average of 18 seeds}} & \multicolumn{4}{c}{\textbf{Best of 18 seeds}} \\ \cmidrule(lr){1-4} \cmidrule(lr){5-8}  \cmidrule(lr){9-12}
    \multirow{2}{*}{\bf Optimizer} & \multirow{2}{*}{\shortstack[c]{\bf Spatial\\ \textbf{Transform}}} & \multirow{2}{*}{\shortstack[c]{\bf Color\\ \textbf{Transform}}} & \multirow{2}{*}{\shortstack[c]{\bf Encoder}} & \multicolumn{2}{c}{\bf Per-pixel} & \multicolumn{2}{c}{\bf LPIPS} & \multicolumn{2}{c}{\bf Per-pixel} & \multicolumn{2}{c}{\bf LPIPS} \\ \cmidrule(lr){5-6} \cmidrule(lr){7-8} \cmidrule(lr){9-10} \cmidrule(lr){11-12}
    & & & &  \textbf{L1} & \textbf{L2} & \textbf{Alex} & \textbf{VGG} & \textbf{L1} & \textbf{L2} & \textbf{Alex} & \textbf{VGG} \\ 
    \midrule
ADAM & & &                                          & 0.98 & 0.62 & 0.41 & 0.58 & 0.83 & 0.47 & 0.33 & 0.51 \\
L-BFGS & & &                                        & 1.04 & 0.68 & 0.45 & 0.61 & 0.85 & 0.49 & 0.35 & 0.53 \\
CMA & & &                                           & 0.96 & 0.61 & 0.39 & 0.55 & 0.91 & 0.54 & 0.37 & 0.54 \\
None & & & \checkmark                               & 1.61 & 1.39 & 0.62 & 0.68 & 1.35 & 1.00 & 0.55 & 0.64 \\ 
\cdashline{1-12}

ADAM & & & \checkmark                               & 0.96 & 0.60 & 0.39 & 0.56 & 0.82 & 0.46 & 0.32 & 0.51 \\
ADAM & & \checkmark &                               & 0.98 & 0.62 & 0.42 & 0.58 & 0.83 & 0.47 & 0.33 & 0.51 \\
ADAM & \checkmark & &                               & 0.90 & 0.54 & 0.44 & 0.57 & 0.76 & 0.41 & 0.36 & 0.50 \\
ADAM & \checkmark & \checkmark & \checkmark         & 0.88 & 0.52 & 0.42 & 0.55 & 0.76 & 0.40 & 0.36 & 0.49 \\ 
\cdashline{1-12}
CMA+ADAM & & &                                      & 0.93 & 0.57 & 0.37 & 0.55 & 0.83 & 0.47 & 0.32 & 0.51 \\
BasinCMA & & &                                      & 0.82 & 0.48 & 0.29 & 0.51 & 0.78 & 0.43 & 0.26 & 0.49      \\
BasinCMA & & & \checkmark                           & 0.82 & 0.47 & \textbf{0.29} & 0.50 & 0.78 & 0.43 & 0.26 & 0.49 \\
BasinCMA & & \checkmark &                           & 0.81 & 0.46 & \textbf{0.29} & 0.50 & 0.77 & 0.42 & \textbf{0.25} & 0.49 \\
BasinCMA & \checkmark & &                           & 0.72 & 0.38 & 0.33 & 0.48 & 0.69 & 0.35 & 0.31 & \textbf{0.46} \\
BasinCMA & \checkmark & \checkmark & \checkmark     & \textbf{0.71} & \textbf{0.37} & 0.32 & \textbf{0.47} & \textbf{0.68} & \textbf{0.34} & 0.31 & \textbf{0.46} \\
\bottomrule
\end{tabular}
}
\medskip
\caption{\small \textbf{ImageNet:} We compare various methods for inverting images from ImageNet (lower is better). The last row is our full method. The model is optimized using $L1$ and AlexNet-LPIPS perceptual loss. The mask and ground-truth class vector is provided for each method. We show the error using different metrics: per-pixel and perceptual~\cite{zhang2018perceptual}. We show the average and the best score among 18 random seeds. Methods that optimized for transformation are inverted to the original location and the loss is computed on the masked region for a fair comparison. All the results here are not fine-tuned.
}
\label{table:score}
\end{center}
\vspace{-0.2in}
\end{table}
\setlength{\tabcolsep}{1.4pt}

\begin{table}[t]
\begin{minipage}[b]{0.49\textwidth}
\centering
\setlength{\tabcolsep}{4pt}
\begin{center}
\resizebox{0.88\linewidth}{!}{
\begin{tabular}{lcccc} 
    \toprule
    & \multicolumn{4}{c}{\textbf{Best of 18 seeds}} \\ \cmidrule(lr){2-5} 
    \multirow{2}{*}{\bf Class search}  & \multicolumn{2}{c}{\bf Per-pixel} & \multicolumn{2}{c}{\bf LPIPS} \\ \cmidrule(lr){2-3} \cmidrule(lr){4-5} 
    & \textbf{L1} & \textbf{L2} & \textbf{Alex} & \textbf{VGG} \\ 
    \midrule
    Random Gaussian            & 1.26 & 0.88 & 0.69 & 0.86 \\
    Random Class               & 0.88 & 0.51 & 0.40 & 0.59 \\
    Predicted                  & 0.84 & 0.47 & 0.33 & 0.52 \\ \cdashline{1-5}
    Ground Truth               & 0.83 & 0.47 & 0.33 & 0.51 \\
\bottomrule
\end{tabular}
}
\vspace{0.05in}
\caption{\small \textbf{Class search:} Given a fixed optimization method (ADAM), we compare different methods for initializing the class vector (lower is better). Baselines are: initialized from $\mathcal{N}(\textbf{0}, \textbf{I}),$ a random class, and the ground truth class.}
\label{table:cls_pred}
\end{center}
\end{minipage}
\hfill
\begin{minipage}[b]{0.49\textwidth}
\centering
\setlength{\tabcolsep}{4pt}
\begin{center}
\resizebox{1.0\linewidth}{!}{
\begin{tabular}{l cccc} 
    \toprule
    & \multicolumn{4}{c}{\textbf{Best of 18 seeds}} \\ \cmidrule(lr){2-5}  
    
    \multirow{2}{*}{\bf Method} & \multicolumn{2}{c}{\bf Per-pixel} & \multicolumn{2}{c}{\bf LPIPS} \\ \cmidrule(lr){2-3} \cmidrule(lr){4-5} 
    & \textbf{L1} & \textbf{L2} & \textbf{Alex} & \textbf{VGG} \\ 
    \midrule
ADAM                    & 0.96 & 0.57 & 0.32 & 0.56 \\
ADAM + Transform        & 0.81 & 0.45 & 0.39 & 0.52 \\
BasinCMA                & 0.93 & 0.18 & \textbf{0.81} & 0.53 \\ 
BasinCMA + Transform    & \textbf{0.78} & \textbf{0.42} & 0.36 & \textbf{0.49} \\
\bottomrule
\end{tabular}
}
\vspace{0.05in}
\caption{\small \textbf{Out-of-distribution:} We compare different methods on the COCO-dataset (lower is better). BigGAN was not trained on COCO images. The class labels are predicted using ResNext-101 and the masks are predicted using MaskRCNN.}
\label{table:coco}
\end{center}
\end{minipage}
\vspace{-0.2in}
\end{table}

\vspace{0.08in}
\xpar{Experimental details.}
We use a learning rate of $0.05$ for $\bz$ and $0.0001$ for $\bc$. We use AlexNet-LPIPS~\cite{krizhevsky2012imagenet,zhang2018perceptual} as our perceptual loss for all our methods. We did observe an improvement using VGG-LPIPS~\cite{simonyan2015very,zhang2018perceptual} but found it to be $1.5$ times slower. In our experiments, we use a total of $18$ seeds for each method. After we project and edit the object, we blend the newly edited object with the original background using Poisson blending~\cite{perez2003poisson}.

For all of our baselines, we optimize both the latent vector $\bz$ and class embedding $\bc$. We use the same mask $\bm$, and the same loss function throughout all of our experiments. 
\ifeccv
    The optimization details of our method and the baselines are in the~Appendix~\red{A}.% \app{app:training}.
\else
    The algorithms we compare against are:
    \vspace{0.03in}
    \begin{itemize}[leftmargin=*,topsep=0pt,itemsep=2pt]%noitemsep,
    
    \item \texttt{ADAM}~\cite{kingma2014adam}: $\bz\sim\mathcal{N}(\textbf{0}, \textbf{I})$ and optimized with ADAM. This method is used in Image2StyleGAN~\cite{abdal2019image2stylegan}.
    
    \item \texttt{L-BFGS}~\cite{liu1989limited}: $\bz\sim\mathcal{N}(\textbf{0}, \textbf{I})$, and optimized using L-BFGS with Wolfe line-search. This method is used in iGAN~\cite{zhu2016generative}.
    
    \item \texttt{CMA~\cite{hansen2001cma}}: $\bz$ is optimized using CMA. 
    
    \item \texttt{ADAM + CMA~\cite{hansen2001cma}}: $\bz$ is drawn from the optimized distribution of CMA. The seeds are further optimized with ADAM.
    
    \item \texttt{ADAM + BasinCMA~\cite{wampler2009basincma}}: $\bz$ is optimized by alternating CMA and ADAM updates (\sect{sec:optimize}).
    
    \item \texttt{+ Encoder~\cite{zhu2016generative,bau2019seeing}}: $\bz$ is initialized with the output of the encoder. To generate variations in seeds, we add a Gaussian noise with a variance of $0.5$. For BasinCMA, the mean of the CMA distribution is initialized with the output of the encoder.
    
    \item \texttt{+ Transform}: Transformation parameter $\phi$ is optimized by alternating CMA updates on $\phi$ and ADAM updates on $\bz$ (\sect{sec:stats}).
    \end{itemize}
    \vspace{0.03in}
    Further optimization details are in the \app{app:training}.
\fi

\begin{figure}[t]
\begin{minipage}[b]{.49\textwidth}
\centering
\includegraphics[width=0.95\linewidth]{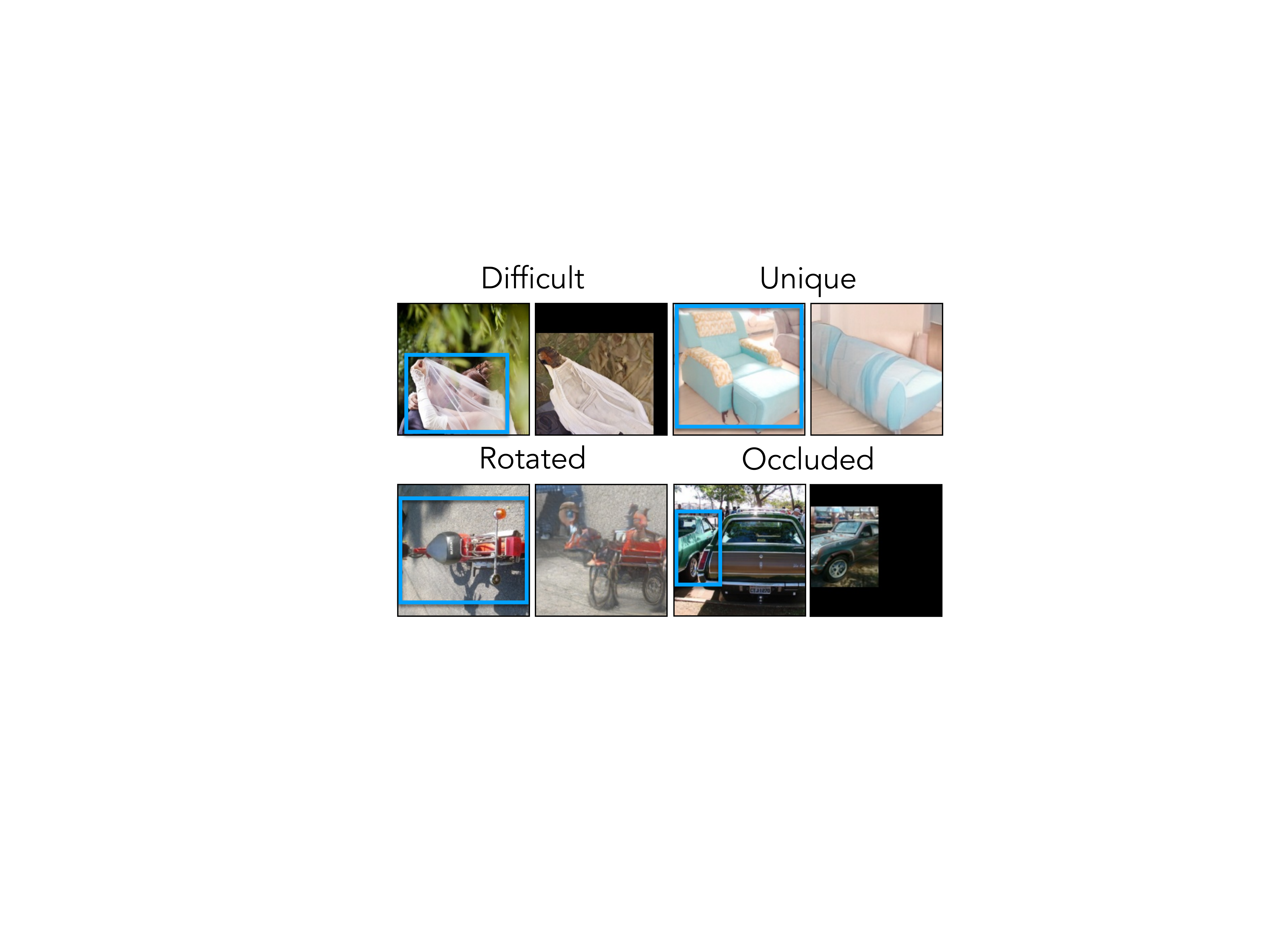}
\vspace{-0.1in}
\caption{\small \textbf{Failure cases:} Our method fails to invert images that are not well represented by BigGAN. The mask is overlayed on the target image in \cyan{blue}.}
\label{fig:failures}
\end{minipage}
\hfill
\begin{minipage}[b]{.49\textwidth}
\centering
\includegraphics[width=0.95\linewidth]{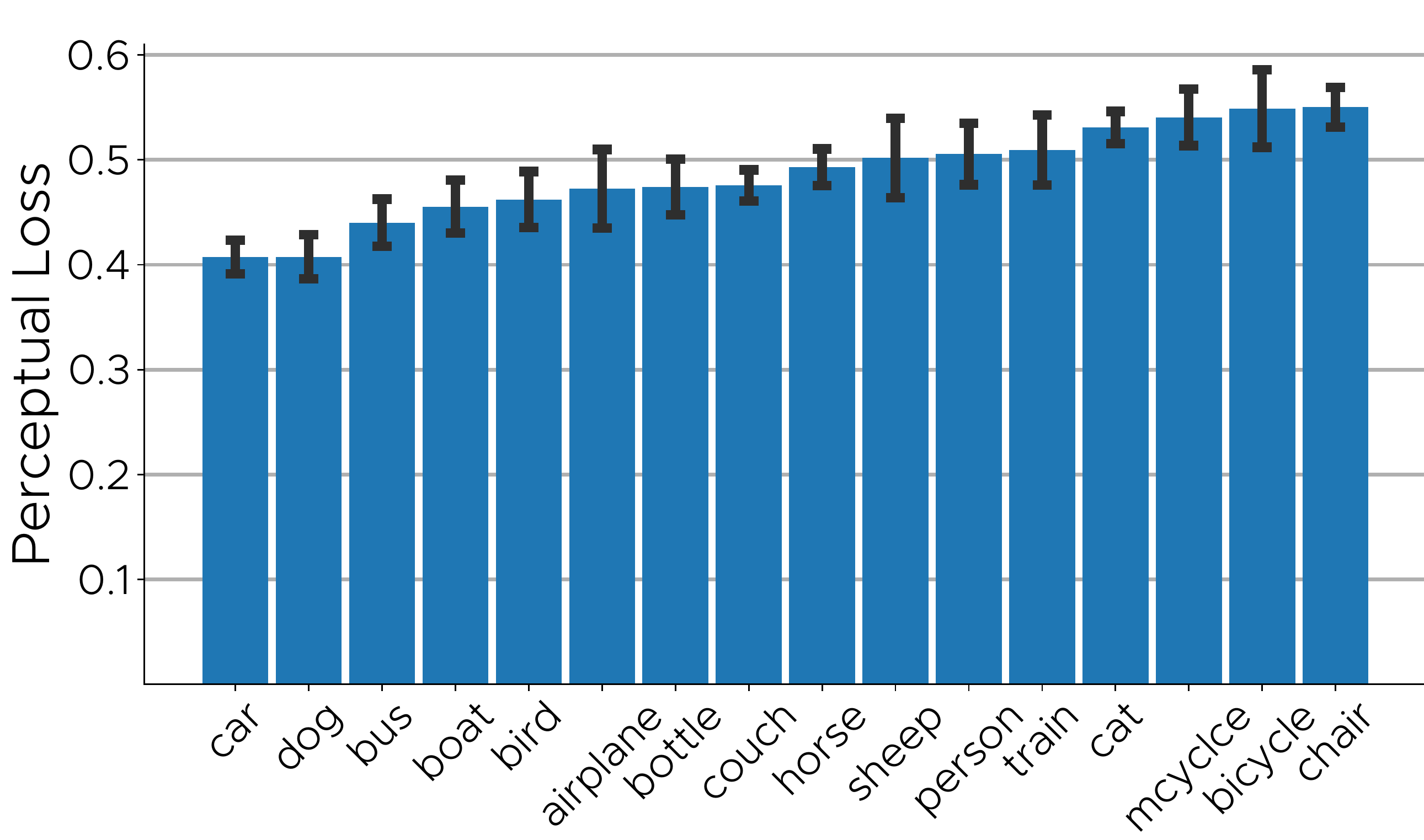}
% \vspace{-in}
\caption{\small \textbf{Projection error by class:} The average VGG-perceptual loss with standard error.
    The ImageNet images are sampled from the PASCAL super-class.}
\label{fig:class_error}
\end{minipage}
\end{figure}

\vspace{0.08in}
\xpar{Experiments.}
We show qualitative comparisons of various optimization methods for ImageNet images in~\fig{fig:comparisons}. We show results of our final method with blending in~\fig{fig:results}.
We then quantify these results by comparing against each method using various metrics in~\tbl{table:score}. For all methods, we do \textit{not} fine-tune our results and we only compute the loss inside the mask for a fair comparison. For methods optimized with transformation, the projected images are inverted back before computing the loss. We further evaluate on COCO dataset~\cite{tsungyi2014coco} in~\tbl{table:coco}, and observed our findings to hold true on out-of-distribution dataset. The success of hybrid optimization over purely gradient-based optimization techniques may indicate that the generative model latent space is locally smooth but not globally. 

\ifeccv
\else
    In~\fig{fig:transform}, we visualize how optimizing for spatial transformation allows us to better fit the target image.
\fi
Without transforming the object, we observed that the optimization often fails to find an approximate solution, specifically when the objects are off-centered or contain multiple objects. 
We observed that optimizing over color transformation does not lead to drastic improvements. Possibly because BigGAN can closely match the color gamut statistics of ImageNet images. Nonetheless, we found that optimizing for color transformation can slightly improve visual aesthetics. Out of the experimented color transformations, optimizing for brightness gave us the best result, and we use this for color transformation throughout our experiments. We further experimented with composing multiple color transformations but did not observe additional improvements.

We found that using CMA/BasinCMA is robust to initialization and is a better optimization technique regardless of whether the transform was applied. Note that we did not observe any benefits of optimizing the class vectors $\bc$ with CMA compared to gradient-based methods, perhaps because the embedding between the continuous class vectors is not necessarily meaningful. Qualitatively, we often found the class embeddings to be meaningful when it is either in the close vicinity of original class embeddings or between the interpolation of 2 similar classes and not more. As a result, we use gradient descent to search within the local neighborhood of the initial class embedding space.

We also provide ablation study on how the number of CMA and ADAM updates for BasinCMA affects performance, and how other gradient-free optimizers compare against CMA in~Appendix~\red{D}. We further provide additional qualitative results for our final method in~Appendix~\red{C}.
% ~\app{app:analysis}
% ~\app{app:add_results}

\vspace{0.08in}
\xpar{Class initialization.}
In downstream editing application, the user may not know the exact ImageNet class the image belongs to. In~\tbl{table:cls_pred}, we compare different strategies for initializing the class vector. Here the classifier makes an incorrect prediction $~20\%$ of the time. We found that using the predicted class of an ImageNet classifier performs almost as well as the ground truth class. Since we optimize the class vector, we can potentially recover from a wrong initial guess if the predicted class is sufficiently close to the ground-truth. 
\begin{figure*}[t!]
    \centering
    \includegraphics[width=1.0\linewidth]{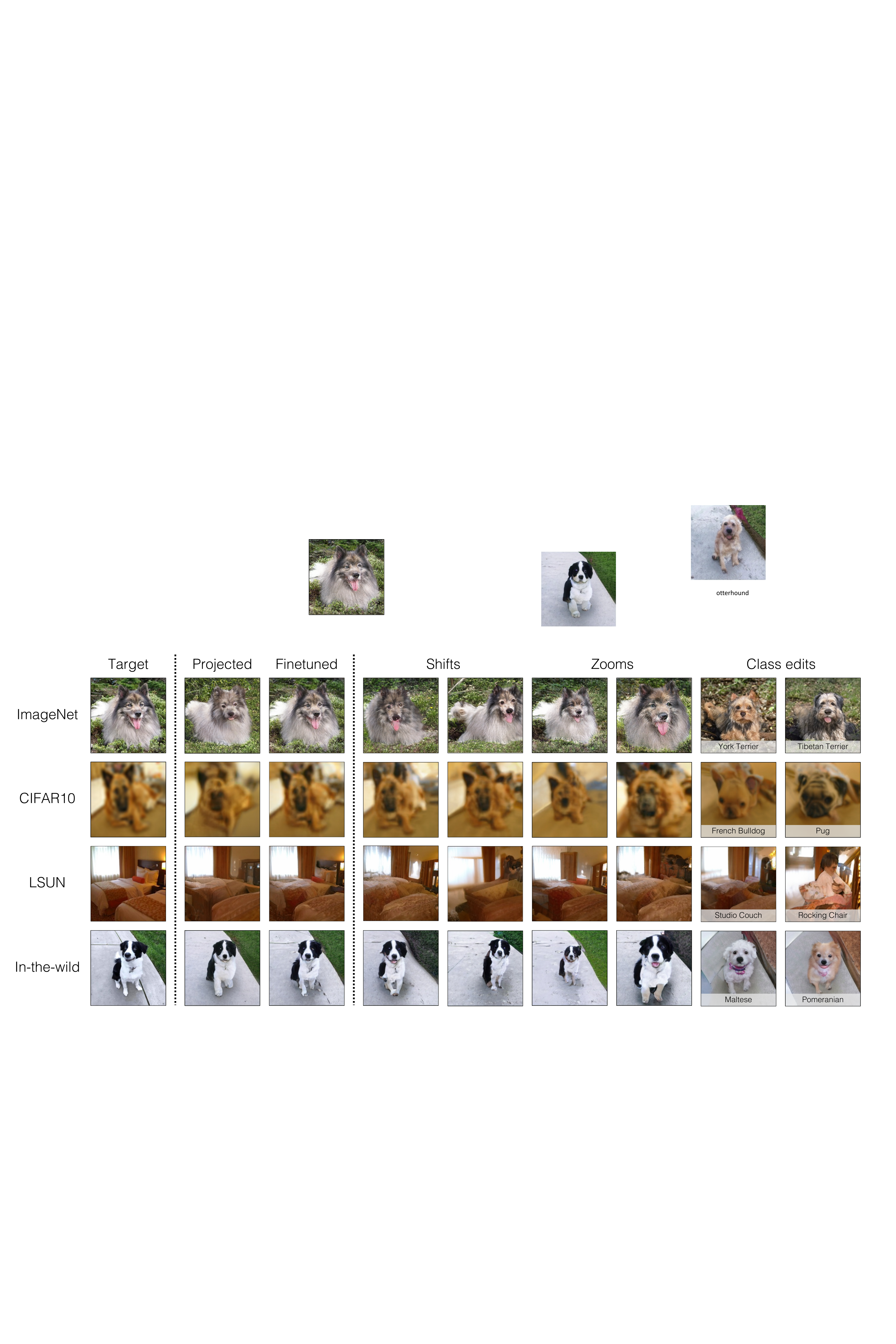}
    \vspace{-0.1in}
    \caption{\small \textbf{Fine-tuned edits:} Inversion results on various datasets. We use BasinCMA and transformation to optimize for the latent variables. After obtaining the projections, we fine-tune the model weights and perform edits in the latent and class vector space.}
    \vspace{-0.2in}
    \label{fig:others}
\end{figure*}

\vspace{0.08in}
\xpar{Failure cases.}
\fig{fig:failures} shows some typical failure cases. We observed that our method fails to embed images that are not well modeled by BigGAN -- outlier modes that may have been dropped. For example, we failed to project images that are unique, complicated, rotated, or heavily occluded. More sophisticated transformations such as rotations and perspective transformation could address many of these failure cases and are left for future work.

\vspace{0.08in}
\xpar{Which classes does BigGAN struggle to generate?}
Given our method, we analyze which classes BigGAN, or our method has difficulty generating. In Figure~\ref{fig:class_error}, we plot the mean and the standard error for each class. The plot is from the output of the method optimized with ADAM + CMA + Transform. We observed a general tendency for the model to struggle in generating objects with delicate structures or with large inter-class variance.

\begin{figure}[t!]
\centering
\includegraphics[width=0.98\linewidth]{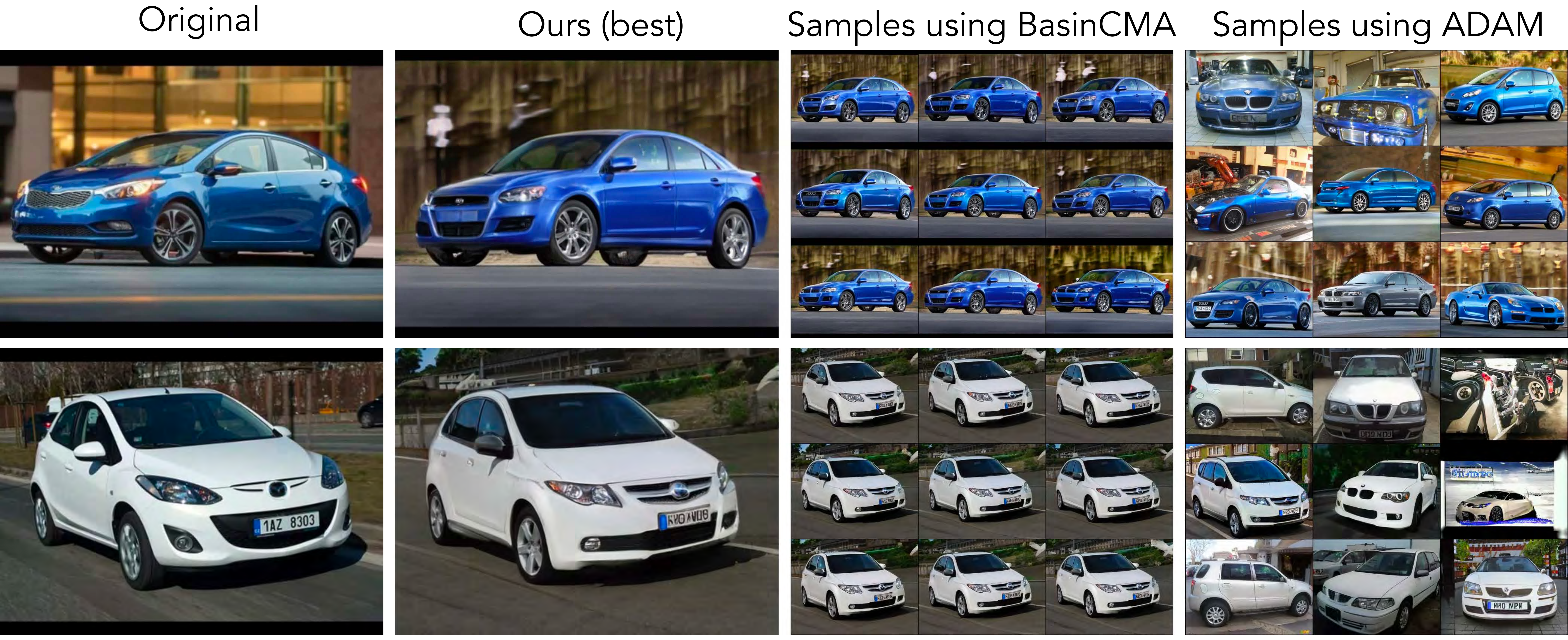}
\includegraphics[width=0.98\linewidth]{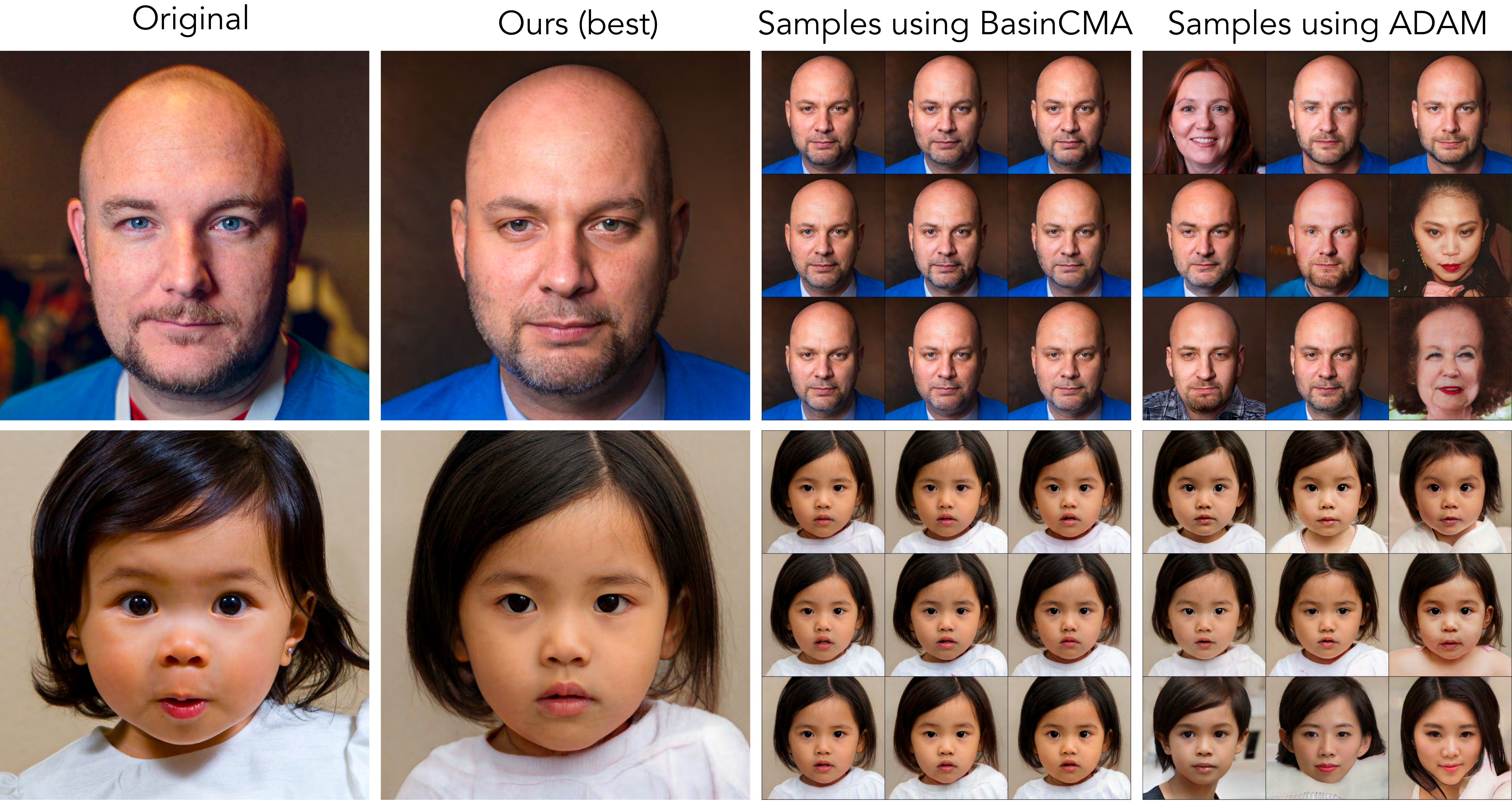}
\vspace{-0.1in}
\caption{\small \textbf{Inverting StyleGAN2 in $z$ space.} We show results of projecting real images into StyleGAN2 using our BasinCMA method without transformation and fine-tuning. The images are inverted into the original input latent code $z\in \mathds{R}^{512}$. The top results are from a model trained on $512\times512$ LSUN cars, and the results on the bottom are from a model trained on $1024\times1024$ FFHQ face dataset. We show results from the top $9$ seeds for both BasinCMA and ADAM.}
\label{fig:stylegan}
\end{figure}

\vspace{0.08in}
\xpar{Image Edits.}
A good approximate solution allows us to fine-tune the generative model and recover the details easily. Good approximations require less fine-tuning and therefore preserve the original generative model editing capabilities. In~\fig{fig:others}, we embed images from various datasets including CIFAR~\cite{krizhevsky2009cifar}, LSUN~\cite{yu15lsun}, and images in-the-wild. We then fine-tune and edit the results by changing the latent vector or class vector. Prior works~\cite{dcgan,jahanian2019steerability} have found that certain latent vectors can consistently control the appearance change of GANs-generated images such as shifting an image horizontally or zooming an image in and out. We used the ``shift'' and ``zoom'' vectors~\cite{jahanian2019steerability} to modify our images. Additionally, we also varied the class vector to a similar class and observed the editability to stay consistent. Even for images like CIFAR, our method was able to find good solutions that allowed us to edit the image. In cases like LSUN, where there is no corresponding class for the scene, we observed that the edits ended up being meaningless. 

\ifeccv
\else
    \vspace{0.08in}
    \xpar{Perceptual Study.}
    We verify the quality of the projected results with a perceptual study on edited projections. We fine-tune each projection to the same reconstruction quality across methods and apply edits to the latent variable $\bz$. We show each image to an Amazon Mechanical Turker for $1$ second and ask whether the edited image is real or fake, similar to~\cite{zhang2016colorful}. Our method (BasinCMA + Transform) achieves $26\%$ marked as real, while the baselines achieve $22\%$ for ADAM, $23\%$ for ADAM + Transform, and $25\%$ for BasinCMA. This indicates that our design choices, adding transforms and choice of optimization algorithm, produces inversions that better enable downstream editing.
\fi

\vspace{0.08in}
\xpar{Inverting unconditional generative model: StyleGAN2.}
StyleGAN~\cite{karras2019style} and StyleGAN2~\cite{Karras2019stylegan2} are other popular choices of generative models for their ability to produce high fidelity images. Although these models can generate high-resolution images, they are restricted to generating images from a single class. Additionally, Abdal et al.~\cite{abdal2019image2stylegan} have demonstrated that it is difficult to project images into the original latent space $z \in \mathds{R}^{512}$ using gradient-descent methods. Henceforth, Image2StyleGAN~\cite{abdal2019image2stylegan} and StyleGAN2~\cite{Karras2019stylegan2} has relied on inverting images into its intermediate representation, also known as the $w^+$~space. The $w^+$~space is $\mathds{R}^{699536}$ for generative model that outputs images of size $512\times512$. Due to the large dimensionality of the intermediate representation, it is much easier to fit any real image into the generative model. Embedding the image into this intermediate representation drastically limits the ability to use the generative model to edit the projected images. On the contrary, we show in~\fig{fig:stylegan} that CMA-based methods can invert the images all the way back to the original latent code $z\in \mathds{R}^{512}$. %and are significantly better alternatives for projecting images into StyleGAN2. 
We observed that models trained on well-aligned images such as FFHQ face dataset~\cite{karras2019style} can often be inverted using gradient-descent methods; however, models trained on more challenging datasets such as LSUN cars~\cite{yu15lsun} can often only be solved using BasinCMA.
\section{Discussion}

Projecting an image into the ``space'' of a generative model is a crucial step %if generative models will be leveraged 
for editing applications. We have systematically explored methods for this projection. We show that using a gradient-free optimizer, CMA, produces higher quality matches. We account for biases in the generative model by enabling spatial and color transformations in the search, and the combination of these techniques finds a closer match and better serves downstream editing pipelines. Future work includes exploring more transformations, such as local geometric changes and global appearance changes, as well as modeling generation of multiple objects or foreground/background.
%
%\ifeccv
%\else
%    

\vspace{4pt}
\xpar{Acknowledgements.}
We thank David Bau, Phillip Isola, Lucy Chai, and Erik H\"ark\"onen for discussions, and David Bau for encoder training code.
%\fi

\clearpage

\bibliographystyle{splncs04}
\bibliography{egbib}

\clearpage
\appendix
% \section*{\centering Appendix for: Transforming and Projecting Images into Class-conditional Generative Networks}

\section{Training and run-time details}
\label{app:training}
The experiments in the appendix were computed on a smaller subset of ImageNet images and are consistent within each other.

For computation, we use a single NVIDIA 2080 TI GPU. The run-time below is with respect to a single GPU. We use a total of $18$ seeds in our main paper. The run-time is an over-estimate as we divide $18$-seeds into $3$ smaller batches to fit it into the GPU memory. 

\vspace{0.08in}
\renewcommand\labelitemi{$\vcenter{\hbox{\tiny$\bullet$}}$}
\begin{itemize}[noitemsep,nolistsep,leftmargin=*]

\item \textbf{ADAM}: $\bz\sim\mathcal{N}(\textbf{0}, \textbf{I})$ and optimized with ADAM. We optimize the latent vector for $500$ iterations, roughly taking $5$ minutes to invert a single image. We observed sharing momentum across random seeds can hurt performance, and we disentangle them in our runs. Furthermore, increasing the number of iterations does not significantly improve performance. This is the optimizer used in Image2StyleGAN~\cite{abdal2019image2stylegan}.
\vspace{0.08in}
\item \textbf{L-BFGS}: $\bz\sim\mathcal{N}(\textbf{0}, \textbf{I})$, and optimized using L-BFGS with Wolfe line-search. We use the PyTorch implementation~\cite{paszke2017automatic} to optimize our latent vector for $500$ iterations. We use the Wolfe line search with an initial learning rate set to $0.1$. L-BFGS has an average run time of $5$ minutes. This is the optimizer used in iGAN~\cite{zhu2016generative}.
\vspace{0.08in}
\item \textbf{Encoder}: We follow the encoder-based initialization methods~\cite{zhu2016generative,bau2019seeing} to train our encoder network on $10$ million generated images, which took roughly $5$~days to train. The encoder network was trained in a class-conditional manner, where the class information was fed into the network through the normalization layers~\cite{vries2017cbn}. We tried using the ImageNet pre-trained model to initialize the weights but found it to perform worse. It takes less than $1$ second to run the encoder but requires additional gradient descent optimization steps for a reasonable result. We observed using an encoder still suffers the same problem as gradient-based methods and slightly improves the results. For our baseline (Encoder + ADAM), we still run ADAM for $500$~iterations.
\vspace{0.08in}
\item \textbf{CMA}: $\bz$ is optimized using CMA. We use the python implementation of CMA~\cite{hansen2019pycma}. For CMA-only optimization, we use $300$ iterations. For CMA+ADAM, we use $100$ CMA iterations and $500$ ADAM updates. It takes roughly $0.2$ seconds per CMA update.
\vspace{0.08in}
\item \textbf{BasinCMA}: $\bz$ is optimized by alternating CMA and ADAM updates. We use the same CMA implementation discussed above with $30$ updates. For each update iteration, we evaluate after taking $30$ gradient steps. The run-time is roughly $10$ minutes per image. Increasing the number of updates and gradient descent steps does improve performance, see~\fig{fig:basincma_ablation}.  
\vspace{0.08in}
\item \textbf{Transformation}: Transformation parameter $\phi$ is optimized by alternating CMA updates on $\phi$ and ADAM updates on $\bz$. We initialize the mean of the CMA using the statistics of generative images, as discussed in~\sect{sec:optimize}. We optimize for $30$ iterations, where CMA is updated after $30$ gradient updates on $\bz, \bc$. Optimizing for transformation adds an additional $5$ minutes.
\vspace{0.08in}
\item \textbf{Encoder}: $\bz$ is initialized with the output of the encoder. To generate variations in seeds, we add a Gaussian noise with a variance of $0.5$. For BasinCMA, the mean of the CMA distribution is initialized with the output of the encoder.
\vspace{0.08in}
\item \textbf{Fine-tuning}: 
    We fine-tune the generative model using ADAM with a learning rate of $10^{-4}$ until the reconstruction loss falls below $0.1$. We use the regularization weight of $10^3$, and we fix the batch-norm statistics during fine-tuning. The whole process takes roughly $1$ minute.
\end{itemize}
\vspace{0.08in}

\begin{figure*}[t!]
    \centering
    \includegraphics[width=0.98\linewidth]{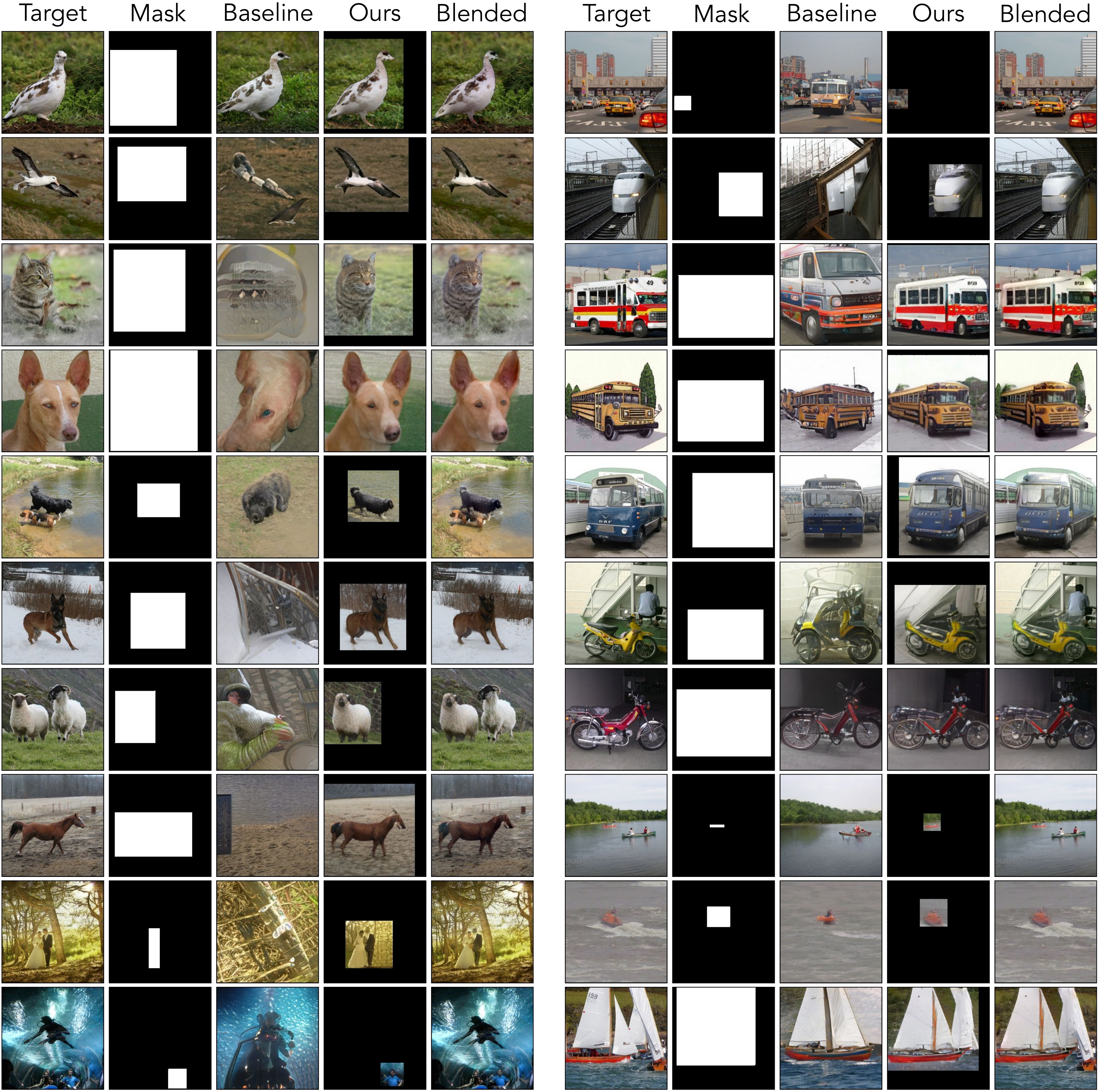}
    \vspace{-.1in}
    \caption{\small \textbf{Additional results:} Comparison between ADAM and our final method. Our method is optimized using BasinCMA, and spatial and color transformation. The results shown above are not fine-tuned.}
    \label{fig:additional}
    \vspace{-.1in}
\end{figure*}

\section{Weighted Perceptual Loss}
\label{app:mLPIPS}
We formulate the weighted LPIPS loss discussed in~\sect{sec:optimize}. Given an input image $\by$, a generated image $\hat{\by}$, we extract the image features from a pre-trained model to compute the loss. The features are extracted from pre-specified $L$ convolutional layers~\cite{zhang2018perceptual}. We denote the intermediate feature extractor for layer $l \in L$ as $F^l(\cdot)$. The features extracted from a real image $\by$ can be written as $F^l(\by) \in \mathbb{R}^{H_l \times W_l \times C_l}$ and similarly for $\hat{\by}$. A feature vector at a particular position is written as $F^l_{hw}(\cdot) \in \mathbb{R}^{C_l}$. LPIPS also provides a per-layer linear weighting $w_l \in \mathbb{R}_+^{C_l}$ to accentuate channels that are more ``perceptual''. To weight the features spatially, we bilinearly resize the mask to match the spatial dimensions of each layer $m^{l} \in [0,1]^{H_l \times W_l}$. Then the spatially weighted loss for LPIPS can be written as:

\begin{equation}
\mathcal{L}_{\text{mLPIPS}}(\by, \hat{\mathbf{y}}, \bm) = \sum_{l \in L} \frac{1}{M^l} \sum_{h,w} m^l_{hw} \lVert w_l \odot (F^l_{hw}(\by) - \hat{F}^l_{hw}(\hat{\by}) ) \rVert_2^2
\end{equation}

Here $\odot$ indicates elementwise multiplication in the channel direction and $M^l$ is the sum of all elements in the mask. 

\section{Additional results}
\label{app:add_results}
We provide additional results for our method without fine-tuning in~\fig{fig:additional}. Our method is optimized using BasinCMA with spatial and color transformation. We provide the ADAM baseline along with our blended result using Poisson blending~\cite{perez2003poisson}.

\begin{figure}[t]
\begin{minipage}[b]{.49\textwidth}
\centering
\includegraphics[width=1.0\linewidth]{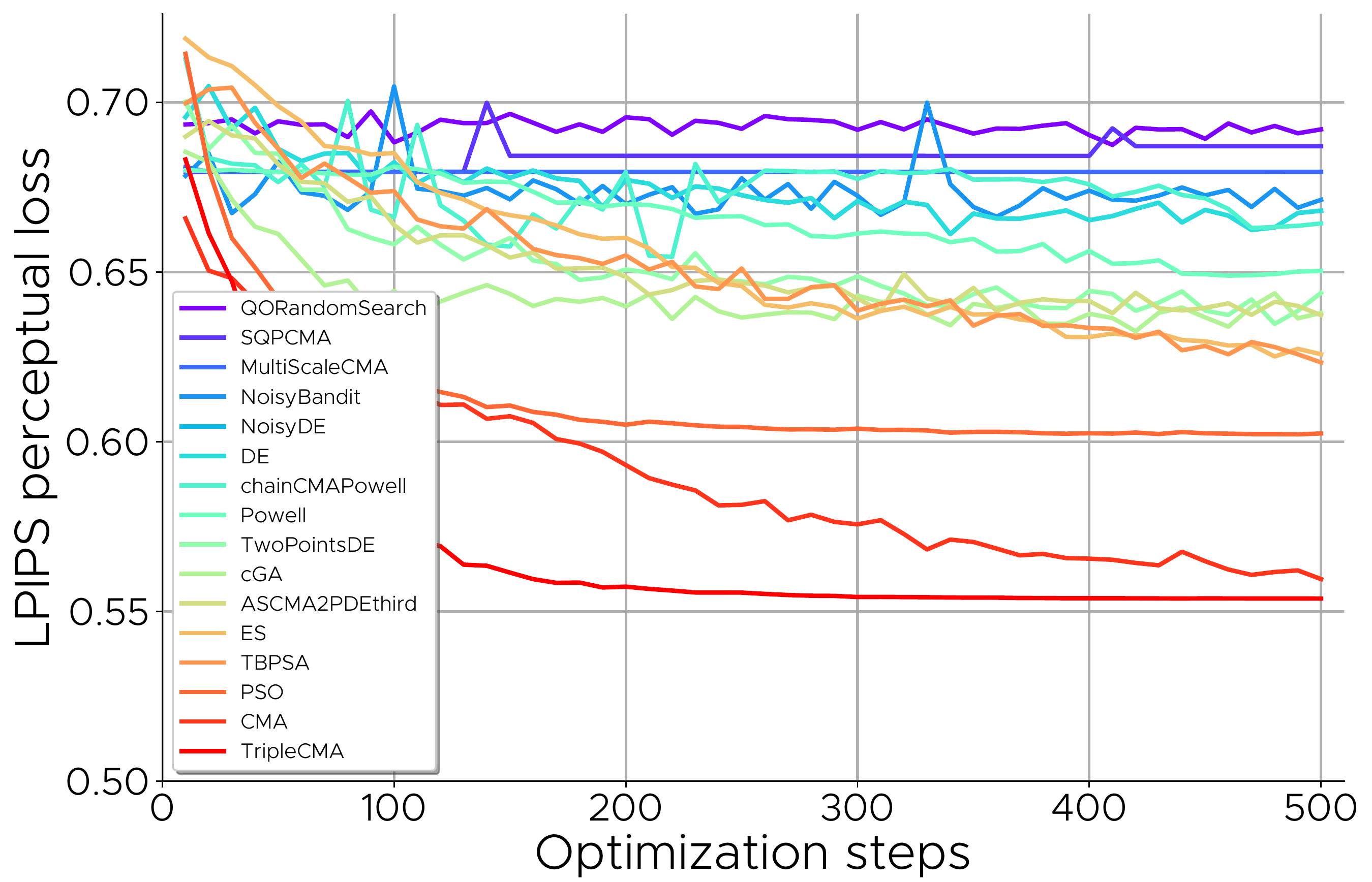}
\vspace{-0.1in}
\caption{\small \textbf{Gradient-free optimizers:} Experiments with various gradient-free optimizers. We use the implementations from Rapin and Teytaud~\cite{nevergrad}. The legend and the color are sorted by performance.}
\label{fig:nevergrad}
\end{minipage}
\hfill
\begin{minipage}[b]{.49\textwidth}
\centering
\includegraphics[width=1.0\linewidth]{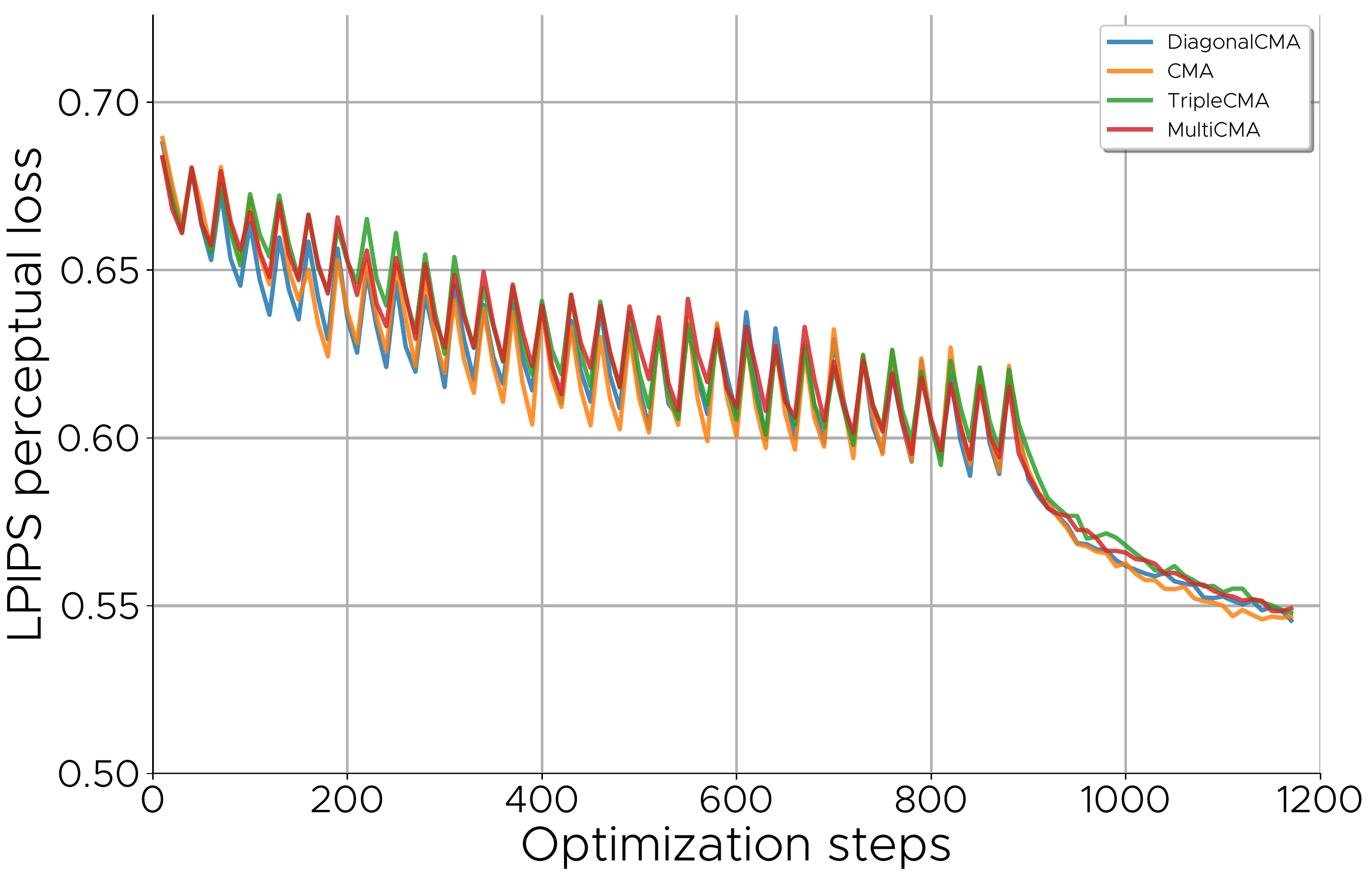}
\vspace{-0.1in}
\caption{\small \textbf{Basin-CMA variants:} Hybrid optimization with different CMA variants. We extended upon the implementations from Rapin and Teytaud~\cite{nevergrad}. All the CMA variants lead to similar results.}
\label{fig:nevergrad_hybrid}
\end{minipage}
\vspace{-0.2in}
\end{figure}

\begin{figure}[t]
\begin{minipage}[b]{.48\textwidth}
\centering
\includegraphics[width=1.0\linewidth]{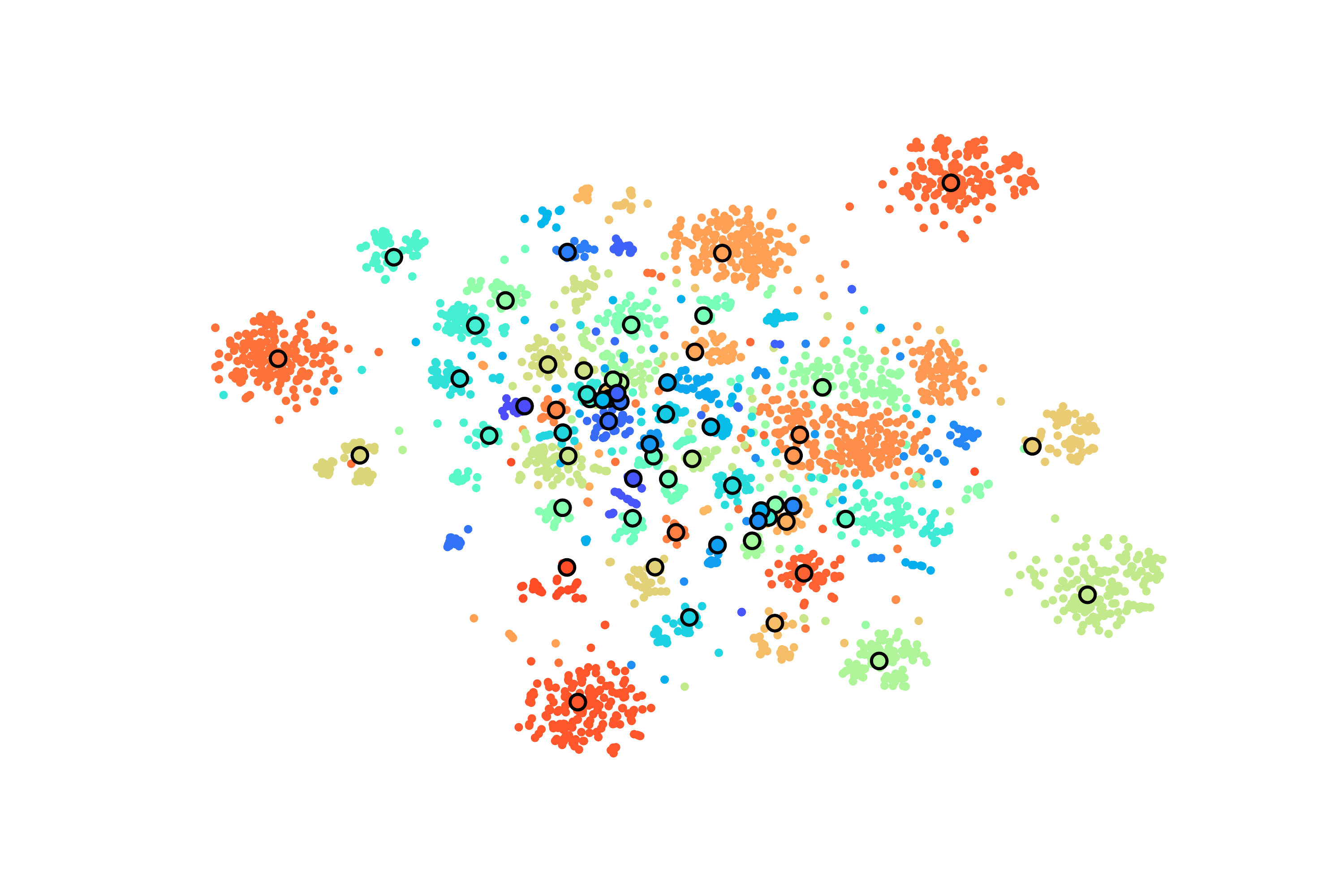}
\vspace{-0.1in}
\caption{\small \textbf{Class vector t-SNE:} The t-SNE embedding of the optimized class vector after optimization. The color represents the class and the circles with black border are the original classes.}
\label{fig:cv_tsne}
\end{minipage}
\hfill
\begin{minipage}[b]{.48\textwidth}
\centering
\includegraphics[width=1.0\linewidth]{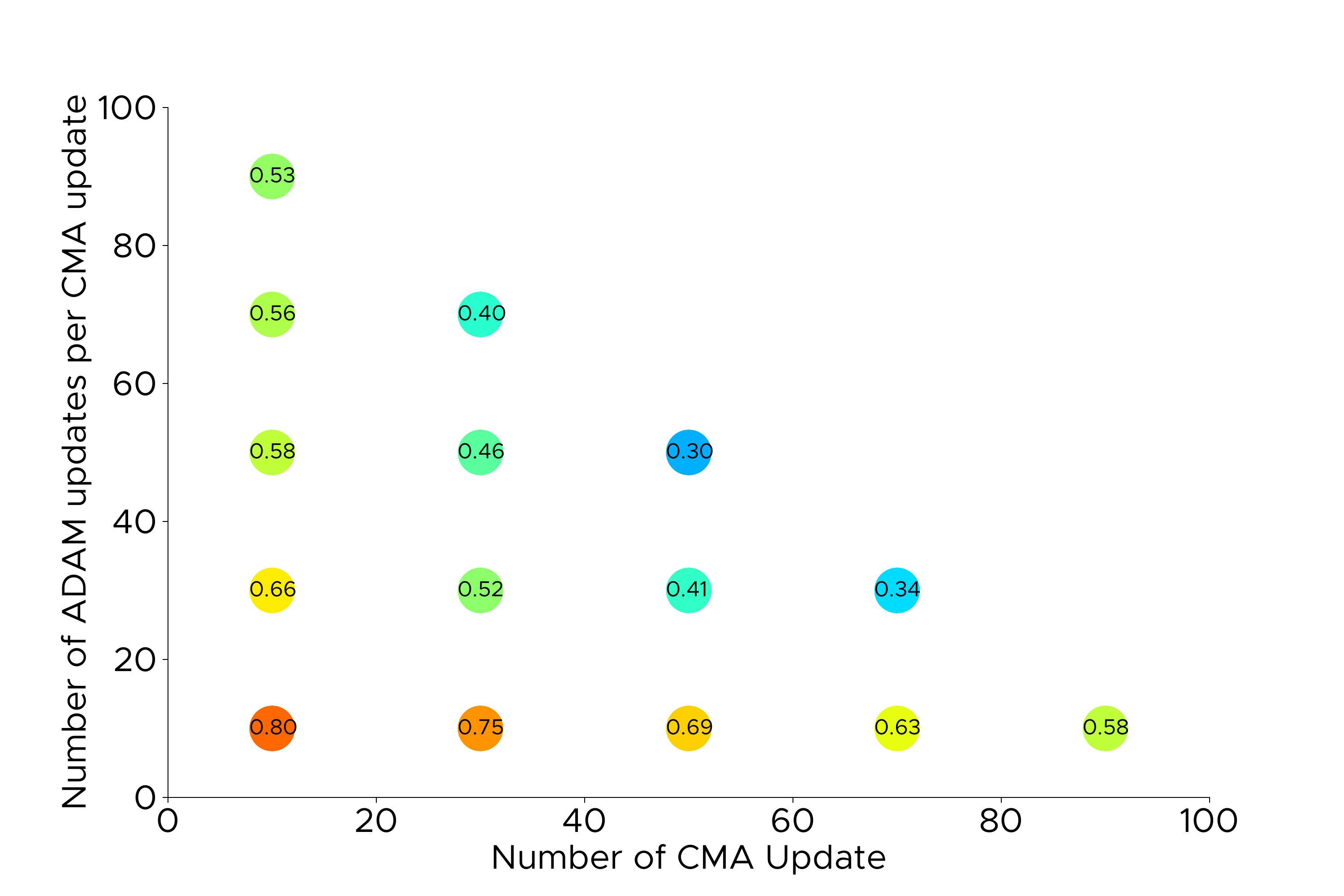}
\vspace{-0.1in}
\caption{\small \textbf{BasinCMA update ablation:} We plot the VGG L-PIPS score when we vary the number of CMA updates (x-axis) and the number of ADAM updates (y-axis). Lower is better. }
\label{fig:basincma_ablation}
\end{minipage}
\vspace{-0.0in}
\end{figure}

\section{Additional Analysis}
\label{app:analysis}

\ifeccv
    \myparagraph{Perceptual study.} We verify the quality of the projected results with a perceptual study on edited projections. We fine-tune each projection to the same reconstruction quality across methods and apply edits to the latent variable $\bz$. We show each image to an Amazon Mechanical Turker for $1$ second and ask whether the edited image is real or fake, similar to~\cite{zhang2016colorful}. Our method (BasinCMA + Transform) achieves $26\%$ marked as real, while the baselines achieve $22\%$ for ADAM, $23\%$ for ADAM + Transform, and $25\%$ for BasinCMA. This indicates that our design choices, adding transforms and choice of optimization algorithm, produces inversions that better enable downstream editing.
\else
\fi

\myparagraph{Inner-outer optimization steps.} Our optimization method maintains a CMA distribution of $\bz$ in the outer loop and is sampled to be optimized in the inner loop with gradient descent. Here the outer loop is the number of CMA updates, and the inner loop is the number of gradient descent updates to be applied before applying the CMA update. In~\fig{fig:basincma_ablation} we ablate the number of optimization steps and observed having the right balance of $1:1$ ratio between CMA and gradient updates leads to the best result. The performance in the figure is mapped by color, with blue indicating the best and red indicating the worst. Although using $50$ CMA update with $50$ ADAM update performs the best, it requires more than $20$-minutes to project a single image. We found $30$ CMA updates and $30$ gradient updates to be a sweet spot for run-time and image quality and is used in all our experiments. We observed the same trend when optimizing for the transformation.

\begin{figure}[t!]
    \centering
    \includegraphics[width=0.9\linewidth]{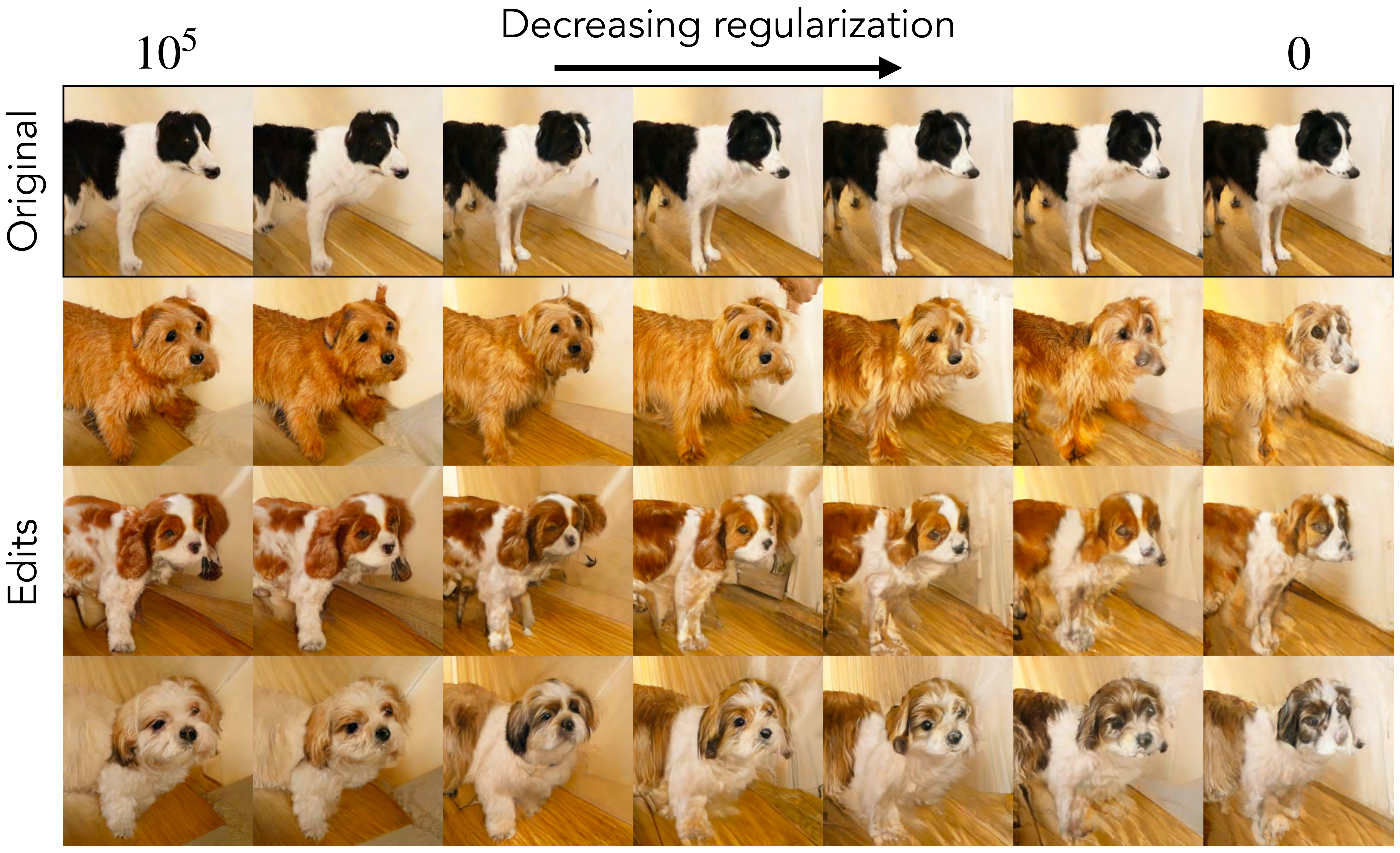}
    \vspace{-.1in}
    \caption{\small \textbf{How fine-tuning affects editing:} We demonstrate how varying the regularization weight effects the edit-ability of the projected image. Images on the top are the original fine-tuned images with varying regularization weight, and the corresponding images below are class edited results. Decreasing the regularization weight allows us to fit the original image better, but introduces more editing artifacts.}
    \vspace{-.2in}
    \label{fig:effect-of-finetune}
\end{figure}

\myparagraph{Speeding up transformation search.} Optimizing for transformation requires the model to quickly search over $\bz$ and $\bc$ given the sampled transformation $F$. Here $\bz$ and $\bc$ are reinitialized at the beginning of the CMA iteration. We observed that initializing $\bz$ by re-using the statistics from the previous iteration can speed up optimization by requiring fewer gradient updates. We found this to be quite important to have the algorithm run efficiently. After thorough testing, we found that sampling from a mutlivariate normal distribution with the mean centered around an exponential moving average of the best performing seeds to work the best. With the decay rate for the exponential moving average set to $0.5$.

\myparagraph{Encoder networks.} Bau et al.~\cite{bau2019semantic} proposed an approach to efficiently train a model-specific encoder $E$ to predict $\bz$ given a generated image $\by$, $E(\by) = \hat{\bz}$. The encoder network is trained only on generated images, and therefore projecting real images often lead to incorrect predictions and require further optimization. Although initializing the optimization with the encoder does not lead to better results, we found that the optimization can converge $20\%$ faster for gradient optimization and $40\%$ faster for hybrid-optimization when $\bz \sim \mathcal{N}(E(\bz), 0.5 \cdot \mathbf{I})$.

\myparagraph{Class-vector embedding.} Optimizing for the class embedding allows the model to better fit the image into the generative model. We provide visualization of optimized class embedding using t-SNE in~\fig{fig:cv_tsne}. The classes are mapped by color and the original classes have black border. We observed that similar classes are embedded closer and class cross-overs are more common during optimization. 

\myparagraph{Gradient-free methods.} We experimented with various gradient-free optimization methods using the Nevergrad library~\cite{nevergrad} in~\fig{fig:nevergrad}. With the default optimization hyper-parameters, we found that CMA and its variants to perform the best. In~\fig{fig:nevergrad_hybrid}, we also experimented with hybrid optimization using various CMA variants but did not see a clear winner.

\myparagraph{How fine-tuning effects editing.} In Figures~\ref{fig:teaser},~\ref{fig:model},~\ref{fig:others}, we demonstrated having good projection allows us to fine-tune the weights to better fit the image without losing the editing capabilities of the generative model. In~\fig{fig:effect-of-finetune}, we visualize how such editing capability is affected by the fine-tuning process. We vary the regularization weight of the fine-tuning objective function that limits the deviation from the original weight. We observed that getting a better initial fit of the image requires us to relax the regularization weight, which in turn introduces additional editing artifacts. Therefore, we found it is crucial to approximate a good initial fit for real image editing.

\section{Changelog}
\myparagraph{v1} Initial preprint release.

\myparagraph{v2} ECCV 2020 camera-ready version. (1) Add related work. (2) Add StyleGAN2 experiments.

\end{document}

%%%%%%%%% TITLE
\maketitle

\section{Training and run-time details}
For computation, we use NVIDIA RTX2080 Ti GPU. The run-time below is with respect to a single GPU and can be significantly sped up by parallelizing across multiple GPUs. We use a total of $9$ seeds in our main paper, the maximum number of seeds that we could fit in a single GPU. Using half-precision improves run-time and memory usage but suffers in projection quality. 

\vspace{0.08in}
\renewcommand\labelitemi{$\vcenter{\hbox{\tiny$\bullet$}}$}
\begin{itemize}[noitemsep,nolistsep,leftmargin=*]
%\textbf{ADAM} \cite{} optimizer used to optimize $\bz$, leaving $\bc$ and $\bt$ fixed;
%\textbf{L-BFGS} \cite{} applied to $\bz$, 
%\textbf{Encoder} \cite{} meaning that the ...
\item \textbf{ADAM}: We optimize the latent vector for $500$ iterations, roughly taking $3$ minutes to invert a single image. We observed sharing momentum across random seeds can hurt performance, and we dis-entangle them in our runs. Furthermore, increasing the number of iterations did not have significant improvement in performance. This is the choice of optimization used in~\cite{abdal2019image2stylegan}.
\vspace{0.08in}
\item \textbf{L-BFGS}: We use the PyTorch implementation~\cite{paszke2017automatic} to optimize our latent vector for $500$ iterations. Instead of using constant step size, we use the Wolfe line search instead. L-BFGS has an average run time of $5$ minutes.
\vspace{0.08in}
\item \textbf{Encoder}: We follow the method~\cite{bau2019seeing} to train our encoder network on $5$ million generated. This took roughly $3$~days to train. The encoder network was trained in a class-conditional manner, where the class information was fed into the network through the normalization layers~\cite{vries2017cbn}. We tried using the ImageNet pre-trained model to initialize the weights but found it to perform worse. It takes less than $1$ second to run the encoder but requires additional gradient descent optimization. We observed using ADAM with the initial prediction of the encoder leads to faster convergence 10-20\% improvement in run-time, but suffers the same problem as gradient methods and does not necessarily lead to better results. For our baseline (Encoder + ADAM) we still run ADAM for $500$~iterations.
\vspace{0.08in}
\item \textbf{CMA}: We use the python implementation of CMA~\cite{hansen2019pycma}. We update the CMA distribution for a fixed $100$ iterations. It takes less than $30$ seconds to run CMA and roughly $4$ minutes when further optimizing with ADAM. We also report experiments where we initialize the mean of the CMA $\mu_0$ with the encoder's prediction (ADAM + Encoder).
\vspace{0.08in}
\item \textbf{BasinCMA}: We use the same CMA implementation discussed above with $30$ updates. For each update iteration, we evaluate after taking $30$ gradient steps. The run-time is roughly $6$ minutes per image with ADAM. Increasing the number of updates and gradient descent steps does slightly improve performance. Reducing the number of updates to couple of iteration still outperforms all the baselines, and henceforth the number of updates can be reduced to minimize optimization run-time.
\vspace{0.08in}
\item \textbf{Transformation}: The transformation search is optimized using CMA. We initialize the mean of the CMA using the generative model statistics discussed in Section~\red{3.4}. We optimize for $50$ iterations, where CMA is updated after taking $10$ gradient updates on the latent vectors $\bz, \bc$. Optimizing for transformation adds an additional $3$ minutes.
\vspace{0.08in}
\item \textbf{Fine-tuning}: 
    We fine-tune the generative model using ADAM with a learning rate of $10^{-4}$ until the reconstruction loss falls below $0.1$. The whole fine-tuning process takes roughly $1$ minute. The batch-norm statistics are held fixed when fine-tuning.
\end{itemize}
\vspace{0.08in}

The run-time above is specific to BigGAN. Note that the optimization speed is directly dependent on the generative model size and perceptual loss. Reducing the computational need for the perceptual loss is an interesting direction which we leave for future work.

\section{Weighted Perceptual Loss}
We provide the weighted L-PIPS loss discussed in section~\red{$3.2$}. Given an input image $\by$, a generated image $\hat{\by}$, we extract the image features from a pre-trained model to compute the loss. The features are extracted from pre-specified $L$ convolutional layers~\cite{zhang_percep}. We denote the intermediate feature extractor for layer $l \in L$ as $\phi_l$. The features extracted from a real image $\by$ can be written as $\phi_l(\by) \in \mathbb{R}^{H_l \times W_l \times C_l}$ and similary for $\hat{\by}$. To weight these features, we bilinearly resize the mask to match the spatial dimensions of each layer $m_{l} \in \mathbb{R}^{H_l \times W_w}$. Then the spatially weighted loss for L-PIPS can be written as:

\begin{equation}
\mathcal{L}_{\text{mLPIPS}}(\by, \hat{\mathbf{y}}, \bm) = \sum_{l \in L} \frac{1}{\lVert m_l \rVert_1} \lVert m_l \odot (\phi_l(\by) - \hat{\phi}_l(\hat{\by}) ) \rVert_1
\end{equation}

Here $\odot$ is the hadamard product, and $\frac{1}{\lVert m_l \rVert}_1$ is a normalization constant computed by taking the sum of all elements in the mask. 

\section{Additional results}
\begin{figure*}[t!]
    \centering
    \includegraphics[width=0.98\linewidth]{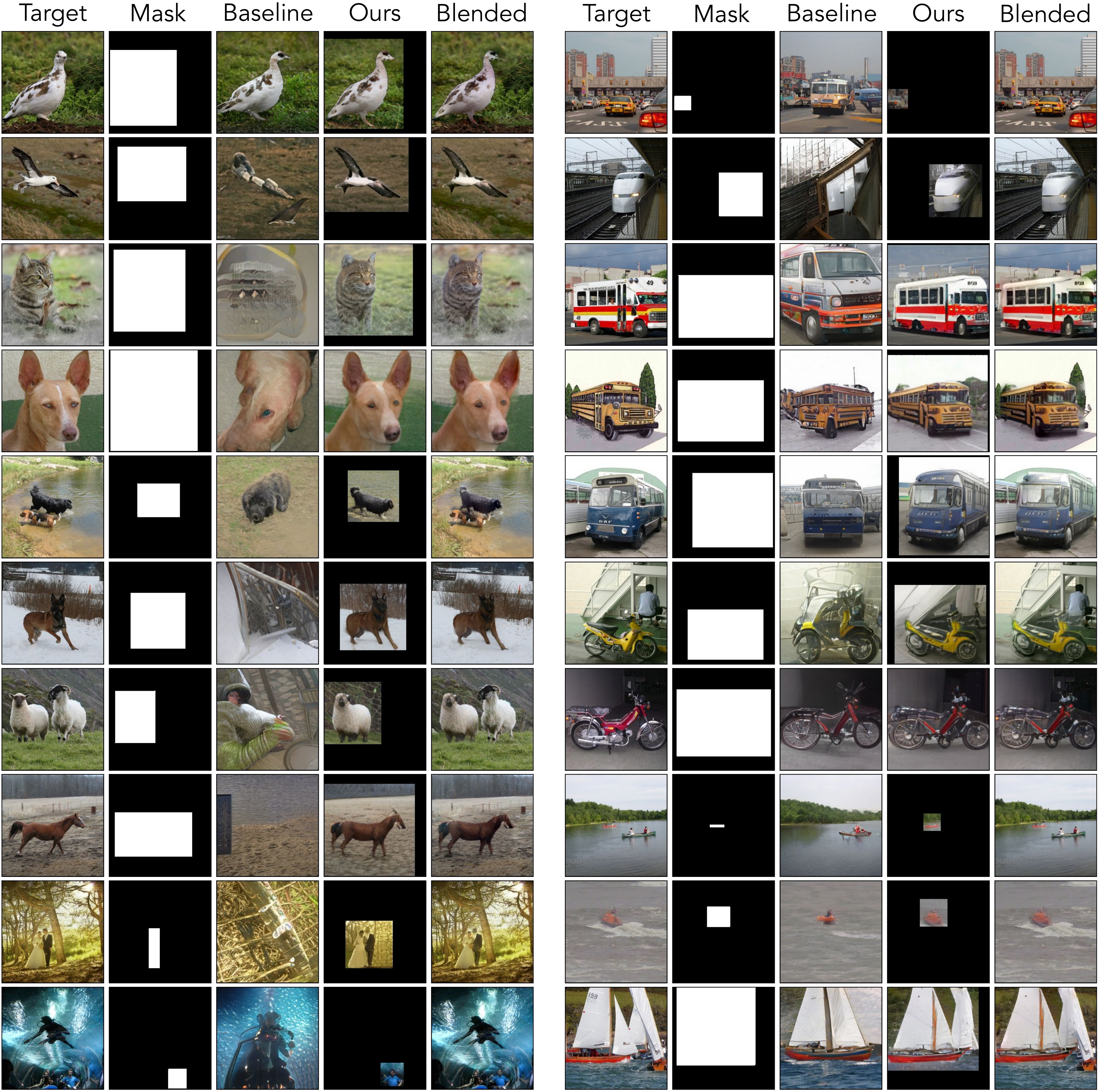}
    \vspace{-.1in}
    \caption{\small \textbf{Additional results:} Comparison between ADAM and our final method. Our method is optimized using BasinCMA, and spatial and color transformation. The results shown above are not fine-tuned.}
    \label{fig:additional}
    \vspace{-.1in}
\end{figure*}

We provide additional results in~\fig{fig:additional} comparing the baseline model~(ADAM) with our method~(BasinCMA + Transform + ADAM). Both methods use the weighted loss, and for our method, the results are projected back into the original location for visualization. Our proposed method outperforms the baseline by finding projections that better resembles the target image. Specifically, for images with multiple objects, we observed the generative model was only able to disambiguate multiple objects when correctly aligned. We further observed that our method can better invert off-centered, occluded, and small object images by searching over transformation. 
%The single-object bias is a limitation of current generative models and may improve over-time. Nonetheless, our proposed method is a model-agnostic and can be easily adapted to future works in generative models.

% Things added later

\section{Ablation studies}
The ablation studies were performed on a smaller subset of ImageNet images and are consistent within each other.

\subsection{Numerical precision}
Generative model such as BigGAN are computationally and memory intensive which becomes the bottle neck of optimization. To this end, we experimented with changing the numerical precision of model weights. We found that lowering the precision by optimizing with half-precision (16-bit) leads to faster wall-clock convergence but results in unstable and sub-par solutions. On the contrary, increasing the numerical precision to double-precision (64-bit) lead to slightly better results but significantly slower optimization speed. Stabilizing the optimization on half precision might be a promising direction for real-time inversion.

\subsection{Inner-outer optimization steps}
Our optimization method maintains a CMA distribution over $\bz$ in the outer loop and are sampled to be optimized in the inner loop with gradient descent. Here the outer loop is the number of CMA updates and the inner loop is the number of gradient descent updates to be applied before applying CMA update. In~\fig{} we ablated the number of optimization steps and found $30$ CMA updates and $30$ gradient updates to be a sweet spot for run-time and image quality. We observed the same trend when optimization for the transformation.

\section{Speeding up transformation search}
Optimizing for transformation requires the model to quickly search over $\bz$ and $\bc$ given the sampled transformation $\phi$. Here $\bz$ and $\bc$ are reinitialized at the beginning of the CMA iteration. We observed that initializing $\bz$ by re-using the statistics from the previous iteration can speed up optimization by requiring less gradient updates. After thorough testing, we found that sampling from a multivariate Gaussian centered around the average of the previous iteration to work the best: $\{\bz_{i}^{t+1}\}_{i=1}^n \sim \mathcal{N}(\frac{1}{n} \sum_{i=1}^n \bz_i^t, 0.3 \cdot \mathbf{I})$, where $\bz_0$ is set to be zero-centered.

\section {Encoder Networks}
Bau et al.~\cite{} proposed an approach to efficiently train a model-specific encoder $E$ to predict $\bz$ given a generated image $\by$, $E(\bz) = \hat{\bz}$. The encoder network is trained only on generated images and therefore projecting real images often lead to incorrect predictions and require further optimization. One might suspect that the encoder can be improved by simply fine-tuning on real images, but the authors noted that doing so actually hurts performance. We found the initial guess of the encoder provides a good starting search space for further optimization: $\bz \sim \mathcal{N}(E(\bz), 0.5 \cdot \mathbf{I})$. Although initializing with the encoder does not lead to better results we found that the optimizer can converge to the same quality $20\%$ faster for gradient optimization and $40\%$ faster for hybrid-optimization. 

\section{Class-vector optimization}
The benefits of optimizing the class embedding are 2 folds: (1) if we have access to the ground-truth class it provides another way to better fit the image into the generative model, (2) if we do not have access to the ground-truth class, we can predict and recover from a wrong initial guess. 

\comm{
    We can optimize for the class embedding by naively optimizing it with gradient descent. This leads to better quantitative numbers but qualitatively worse results. We also found that editing the image is harder when the class embedding is optimized. To visualize the results, we embedded the optimized class embeddings using t-SNE in~\fig{} and observed significant drifting from the ground truth embedding; a potential cause of poor edit-ability and qualitative result. We quantified this drift from the ground-truth class embedding in~\fig{} and observed an almost linear increase in L2 distance from the ground truth class embedding.
    
    To address this issues, we limit the search space of the class embedding by optimizing over the linear combination coefficients of potential class candidates. The class embedding can be constructed as:
    \begin{align}
    \bc_{lin} = \sum_{k=1}^{K} \text{softmax}(\alpha_k, \alpha_{1:K}, \tau) \bc_{k} 
    \end{align}
    Here $K$ is the number of candidates and $\bc_k$ is the corresponding candidate class embedding and $\alpha_k$ is the un-normalized scalar coefficient weight. The candidates and the coefficients are initialized using the top-$K$ prediction of the classifier and it's corresponding logits. The coefficients are normalized using the softmax operator with the temperature set to $\tau=3$. We use $K=5$ for all our experiments. As shown in~\fig{}, the embedding stays significantly close to the ground-truth class-vector. Furthermore, we observed that initializing the coefficients with incorrect predictions or with uniform probability can still quickly snap on to the correct class as long as there exists a candidate class that is sufficiently close to the ground-truth class.  
}
\begin{figure}[t]
\begin{minipage}[b]{.48\textwidth}
\centering
\includegraphics[width=1.0\linewidth]{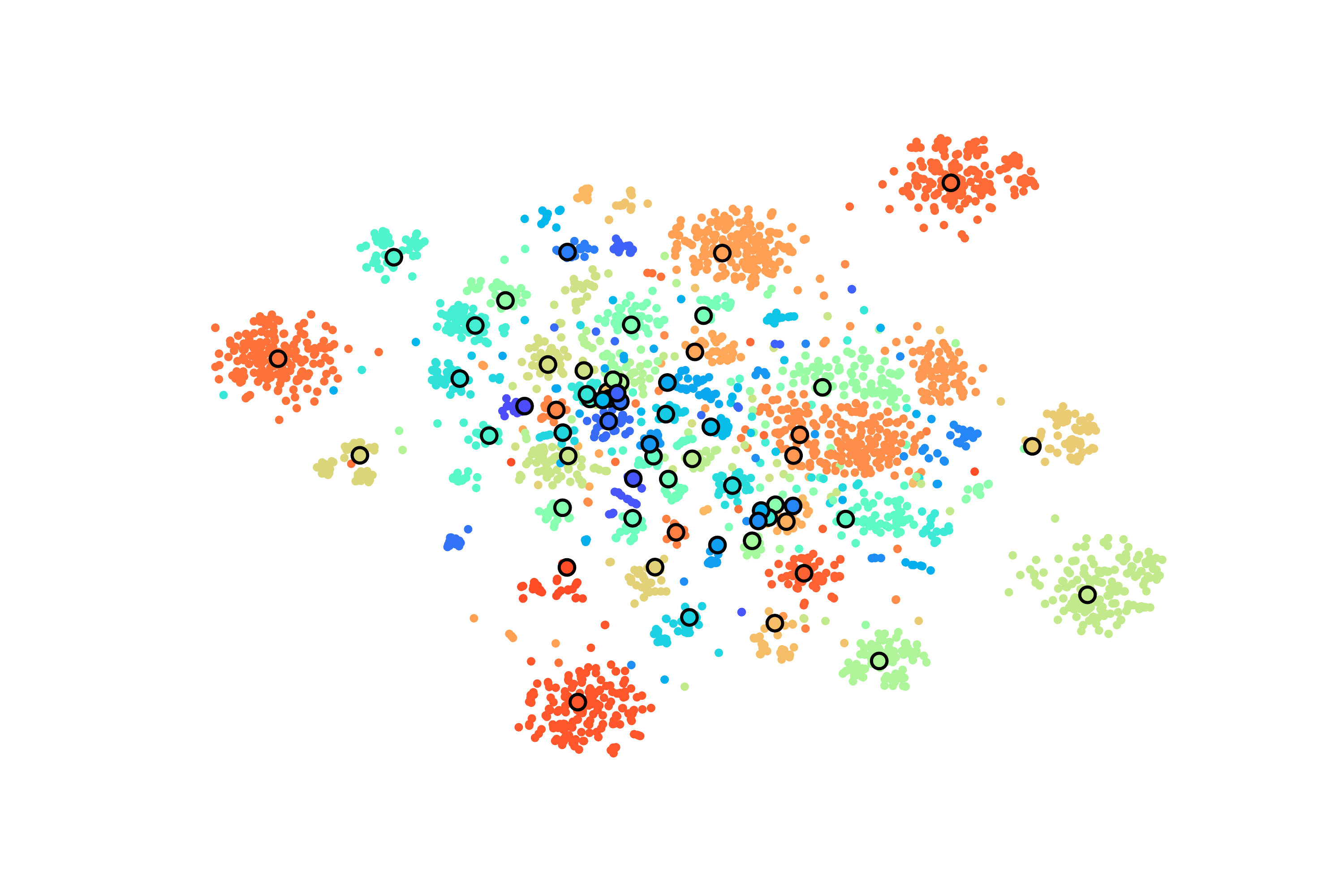}
\vspace{-0.1in}
\caption{\small \textbf{Class vector t-SNE:} The t-SNE embedding of the optimized class vector after optimization. The color represents the class and the circles with black border are the original classes.}
\label{fig:cv_tsne}
\end{minipage}
\hfill
\begin{minipage}[b]{.48\textwidth}
\centering
\includegraphics[width=1.0\linewidth]{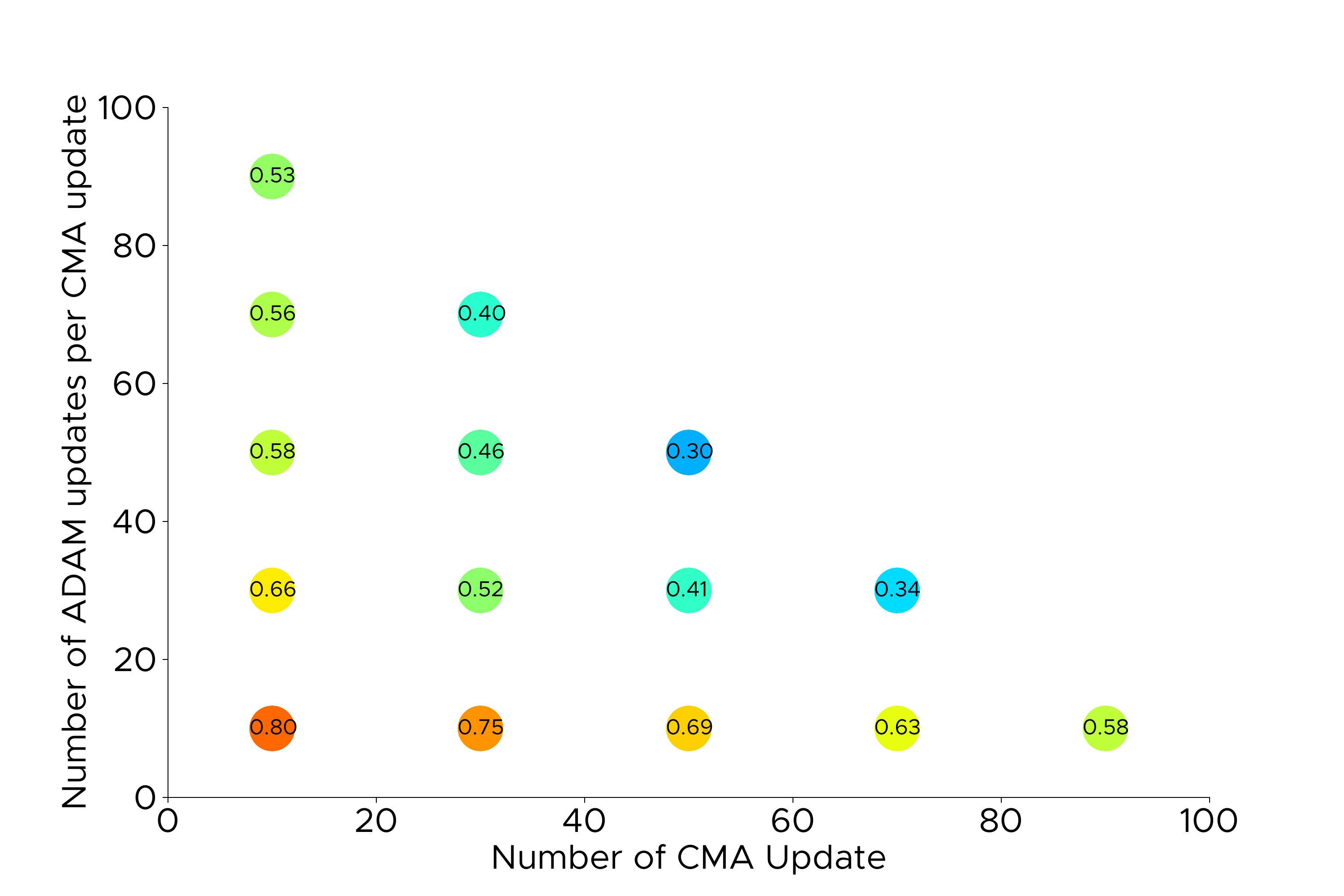}
\vspace{-0.1in}
\caption{\small \textbf{BasinCMA update ablation:} We plot the VGG L-PIPS score when we vary the number of CMA updates (x-axis) and the number of ADAM updates (y-axis). Lower is better. }
\label{fig:basincma_ablation}
\end{minipage}
\vspace{-0.0in}
\end{figure}

\begin{figure}[t]
\begin{minipage}[b]{.49\textwidth}
\centering
\includegraphics[width=1.0\linewidth]{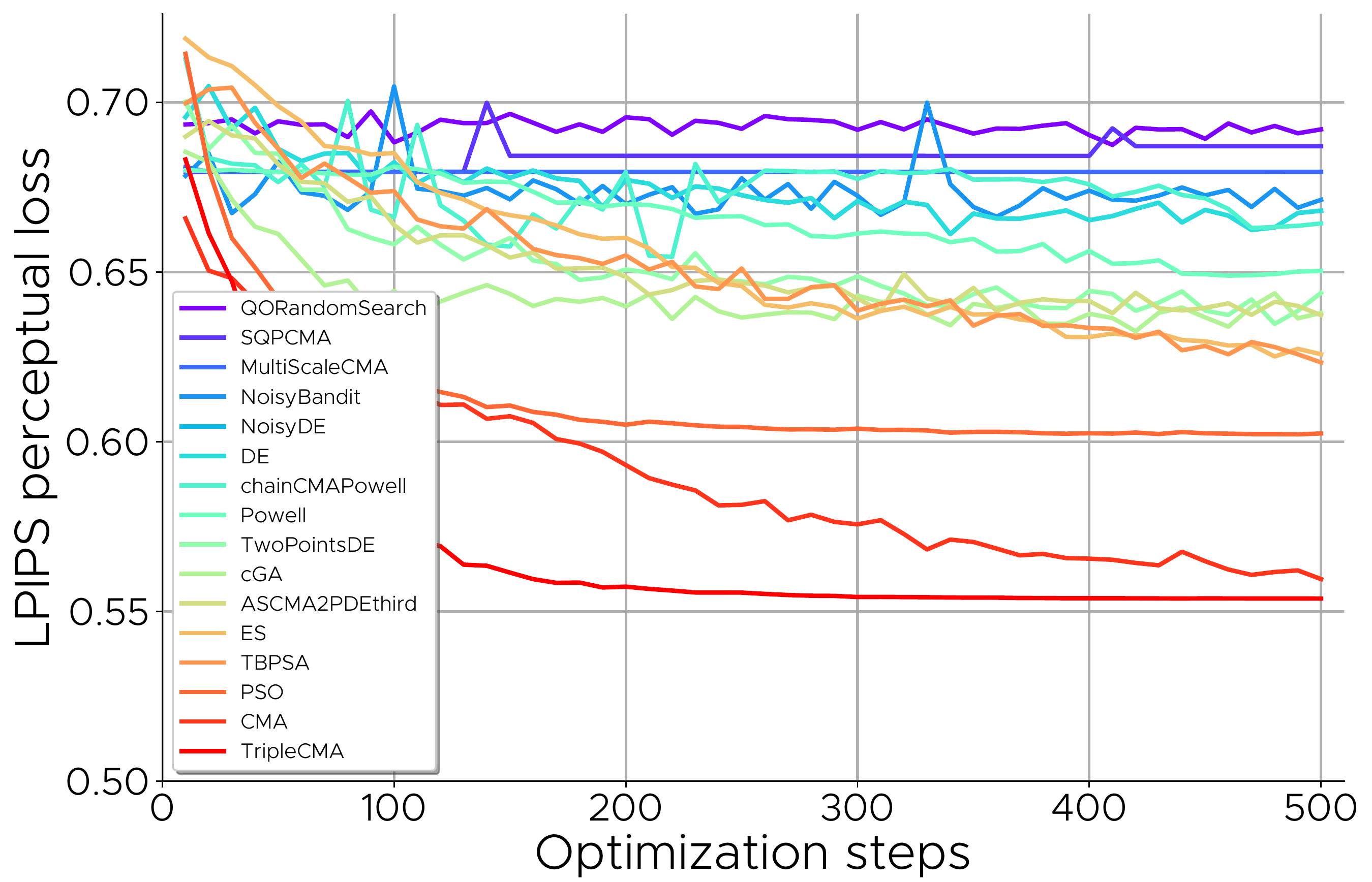}
\vspace{-0.1in}
\caption{\small \textbf{Gradient-free optimizers:} Experiments with various gradient-free optimizers. We use the implementations from Rapin and Teytaud~\cite{nevergrad}. The legend and the color are sorted by performance.}
\label{fig:nevergrad}
\end{minipage}
\hfill
\begin{minipage}[b]{.49\textwidth}
\centering
\includegraphics[width=1.0\linewidth]{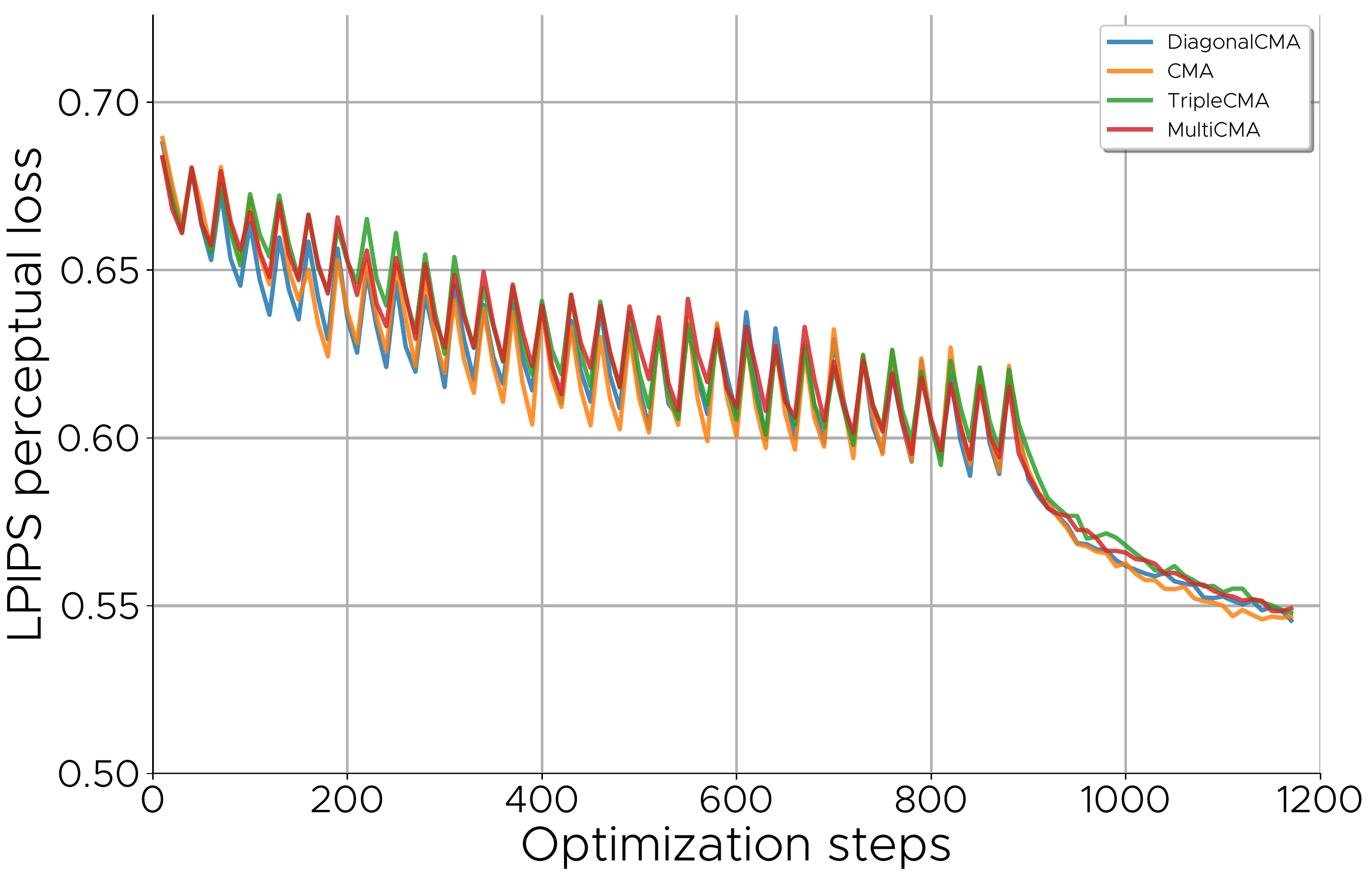}
\vspace{-0.1in}
\caption{\small \textbf{Basin-CMA variants:} Hybrid optimization with different CMA variants. We extended upon the implementations from Rapin and Teytaud~\cite{nevergrad}. All the CMA variants lead to similar results.}
\label{fig:nevergrad_hybrid}
\end{minipage}
\vspace{-0.2in}
\end{figure}

\bibliographystyle{splncs04}
\bibliography{egbib}